\documentclass{ecai} 

\usepackage{times}
\usepackage{soul}
\usepackage{url}
\usepackage{graphicx}
\usepackage{amsmath}
\usepackage{amsthm}
\usepackage{booktabs}
\usepackage{etoolbox}
\usepackage[switch]{lineno}
\usepackage{natbib}
\usepackage{amssymb}
\usepackage{mathabx}
\usepackage{comment}
\usepackage{multirow}
\usepackage{enumitem}
\usepackage{color}
\usepackage{subcaption}
\usepackage{algorithm}
\usepackage{algorithmic}
\usepackage{placeins}

\newcommand{\ENDIFadvice}[1]{\State \hlADVICE{\textbf{end if}}}

\newcommand{\ENDFORadvice}[1]{\State \hlADVICE{\textbf{end for}}}

\usepackage{xcolor}
\usepackage{fancyvrb}
\usepackage{tcolorbox}

\colorlet{myorange}{orange!40}

\newcommand{\changed}[1]{\textcolor{black}{#1}}
\newcommand{\approach}{ADVICE}

\usepackage[capitalize,noabbrev]{cleveref}

\theoremstyle{plain}
\newtheorem{theorem}{Theorem}

\theoremstyle{definition}

\theoremstyle{remark}

\begin{document}


\begin{frontmatter}


\paperid{4326} 


\title{Safe Reinforcement Learning in Black-Box Environments via Adaptive Shielding}


\author[A]{\fnms{Daniel}~\snm{Bethell}
\thanks{Corresponding Author. Emails: firstname.lastname@york.ac,uk}
}
\author[A,B]{\fnms{Simos}~\snm{Gerasimou}}
\author[A]{\fnms{Radu}~\snm{Calinescu}} 
\author[A]{\fnms{Calum}~\snm{Imrie}} 

\address[A]{University of York, UK}
\address[B]{Cyprus University of Technology, Cyprus}


\begin{abstract}
Safe exploration of reinforcement learning (RL) agents is a critical activity for empowering their deployment in many real-world scenarios. When prior knowledge of the target domain or task is unavailable, training RL agents in unknown, \textit{black-box} environments unavoidably yields significant safety risks. 
Our \mbox{ADVICE} (Adaptive Shielding with a Contrastive Autoencoder) novel post-shielding approach operates in continuous state and action spaces, distinguishing safe and unsafe features of state-action pairs during training, and uses this knowledge to safeguard the RL agent from executing actions that yield likely hazardous outcomes. 
Our comprehensive experimental evaluation shows that ADVICE significantly reduces safety violations ($\approx\!\!50\%$) compared to state-of-the-art safe RL exploration approaches, while maintaining a competitive outcome reward for the synthesised safe policy.
\end{abstract}

\end{frontmatter}


\section{Introduction}
Reinforcement Learning (RL)~\citep{sutton2018reinforcement} is a powerful machine learning paradigm for solving complex decision-making tasks. RL has exhibited performance commensurate with the cognitive abilities of humans in diverse applications, including game-playing~\citep{open-ai-five} and robot control~\citep{massively-drl, loc-behaviours}. Despite its huge potential, developing RL-based agents that can explore their environment safely remains a major challenge. Exploring unfamiliar, and potentially hazardous, states while learning from the environment, poses real dangers especially in safety-critical domains, like robotics or healthcare. 
Alleviating this challenge requires RL agents that can safely explore their environment by avoiding unsafe actions while simultaneously devising an optimal policy by exploring the policy space adequately~\citep{concrete-problems}.

Ensuring safety becomes an increasingly difficult challenge as environments characterised by continuous state/action spaces become progressively more complex~\citep{dalal2018safe}. Such environments require a large amount of training time before the RL agent can consistently complete the task and avoid safety concerns. This issue is further exacerbated in \textit{black-box} environments where prior or domain knowledge is not available in advance; only data observed in real-time by the RL agent is available. In such scenarios, the risk associated with exploration increases exponentially as the agent must operate without pre-defined guidelines or guardrails, rendering typical safe exploration approaches inadequate~\citep{waga2022dynamic}. 

Prior research on safe RL exploration formulates safety constraints as linear temporal~\citep{konighofer2023online} and probabilistic logic~\citep{yang2023safe} specifications, whose use as a shield protects the agent during training. 
Shielding approaches are categorised into pre-shielding (restricting action choices to a predefined safe subset) and post-shielding (evaluating and modifying actions post-selection to ensure safety)~\citep{shieldreview}. 
Despite major advances, a common challenge across these approaches is their reliance on some degree of prior knowledge about the environment, task, or safety concern---a strong assumption that is not generally applicable~\citep{waga2022dynamic}.
Research targeting safe exploration in black-box environments employs Lagrangian approaches~\citep{achiam2017constrained,stooke2020responsive}, or involves a pre-training phase before the shield synthesis~\citep{tappler2022automata}. However, these approaches require an explicit cost signal at every timestep or a small amount of prior knowledge to initialize the shield, neither of which is available in many environments.

\begin{figure*}[tb!]
    \centering
    \includegraphics[width=\linewidth]{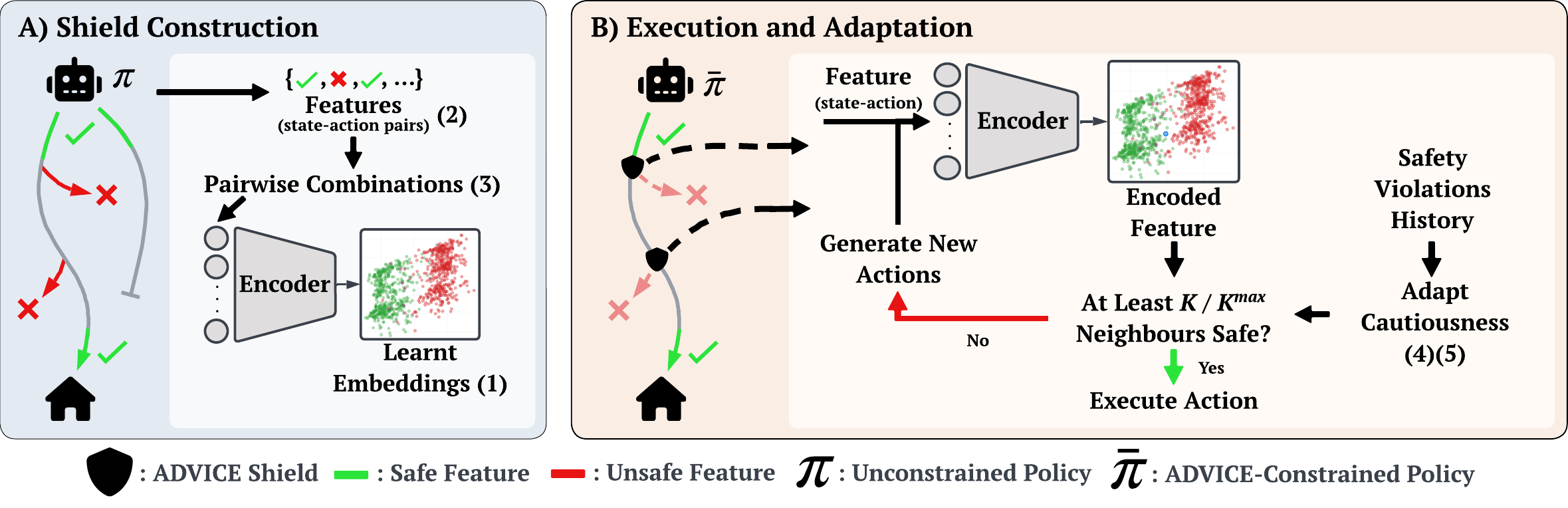}
    \caption{High-level ADVICE overview, showing its shield construction (A), and execution and adaptation (B) stages. During shield construction, safe and unsafe features (signifying accepting and non-accepting terminal states, respectively) are collected and used by a contrastive autoencoder to learn latent representations (embeddings). During shield execution and adaptation, a new feature is encoded into the contrastive latent space and the nearest neighbours 
    guide its classification as safe or unsafe. 
    ADVICE's cautiousness is adapted dynamically based on the RL agent's recent safety performance. A detailed ADVICE description is provided in Section~\ref{sec:advice}.}
    \label{fig:advice-overview-large}
    \vspace{2mm}
\end{figure*}

We mitigate this significant limitation by introducing ADVICE (\textbf{AD}apti\textbf{V}e Sh\textbf{I}elding with a \textbf{C}ontrastive Auto\textbf{E}ncoder), a novel post-shielding approach for safe RL agent exploration in \textit{black-box} environments with continuous state and action spaces, where no prior knowledge of the environment dynamics, and reward structure is available. Following \cite{alshiekh2018safe, belardinelli2025probabilistic}, \approach\ relaxes traditional cost-based safety assumptions by operating within the standard MDP framework, defining safety through accepting and non-accepting terminal states, thereby enabling greater generalisability across tasks and environments without requiring explicit cost signals. Unlike many prior shielding approaches, ADVICE does not rely on handcrafted domain knowledge or expert-defined safety constraints, enabling full end-to-end shield synthesis in black-box environments. To the best of our knowledge, ADVICE is the first shielding approach for safe RL in black-box environments with continuous state/action spaces, offering an end-to-end shield synthesis without using \emph{any} prior knowledge.
Figure~\ref{fig:advice-overview-large} shows a high-level overview of \approach's two-stage approach. During shield construction (Figure~\ref{fig:advice-overview-large}A), ADVICE uses a contrastive autoencoder to learn latent representations that distinguish safe and unsafe \emph{features} (i.e., state-action pairs). During execution (Figure~\ref{fig:advice-overview-large}B), ADVICE employs the learnt representations and a non-parametric classifier to identify and correct unsafe actions. ADVICE also dynamically adapts its cautiousness based on the agent's recent performance, encouraging exploration when suitable. Finally, ADVICE demonstrates effectiveness in complex \textit{black-box} environments, significantly reducing safety violations compared to state-of-the-art approaches like Lagrangian~\citep{ddpg-paper} and discretized shields~\citep{tabular-shield}, and conservative safety critics~\citep{csc-agents}.


\section{Related Work}
\label{sec:rel-work}

\textbf{Shielding Approaches.} 
Recent research has devised approaches for the safe exploration of RL agents using safety shields, allowing the agent to select from a pool of safe actions or correct an action deemed unsafe~\citep{shieldreview}. Existing shielding approaches leverage linear temporal logic (LTL) specifications~\citep{alshiekh2018safe, elsayed2021safe,konighofer2023online} or use external hints for constructing the LTL formulae~\citep{waga2022dynamic}. LTL specifications can be replaced with probabilistic logic programming~\citep{yang2023safe}, extending their use to continuous deep RL and enabling safety constraints to be differentiable. By utilising logical neural networks~\citep{kimura2021reinforcement}, the same logic specifications can be both respected and learnt, providing a more nuanced understanding of safety. \citet{jansen2020safe} introduce probabilistic shields to ensure safety, 
while~\citet{carr2023safe} implement safety under partial observability and~\citet{goodall2023approximate} use approximate models of the environment to maintain safety. \citet{dalal2018safe} train a safety layer to adhere to a known constraint function.
Although these approaches can improve safety and, in limited cases, eliminate violations, their demand for explicit prior knowledge of the RL agent environment, task, and/or safety concerns, restricts
their applicability to a narrow set of environments and safety aspects~\citep{turchetta2020safe}. 
In contrast, our \approach\ post-shielding approach does not need \emph{any} handcrafted prior knowledge, and uses exclusively the information captured in a typical RL problem. 

\textbf{Black-Box Safe Exploration Approaches.} Other recent research focuses on improving safety in \textit{black-box} environments, where no prior knowledge is provided to the agent/user. A trivial but effective solution is to record all unsafe features in a tabular format to prevent the agent from repeating them~\citep{tabular-shield}. However, the applicability of this approach is limited to discrete environments or extremely low-dimensional spaces. Other solutions collect environment data before training, to then instrument a safety layer~\citep{srinivasan2020learning, csc-agents} or shield~\citep{tappler2022automata} that protects the agent. This requires a significant amount of data collection, resulting in a prohibitive number of safety violations. Lagrangian approaches enable modelling the Markov Decision Process, encoding the RL problem as a Constrained Markov Decision Process, and are widely adopted due to their simplicity and effectiveness~\citep{altman2021constrained, garcia2015comprehensive}. The Lagrangian multiplier can be fixed~\citep{altman1998constrained,stooke2020responsive}, or integrated into the algorithm itself~\citep{achiam2017constrained,tessler2018reward}. Other solutions employ uncertainty estimation concepts~\citep{jain2021safe,kahn2017uncertainty}. Defining safety in terms of uncertainty and propagating it into the RL algorithm during training yields a cautious yet effective agent for reducing safety violations, even in complex environments. Similarly to these approaches, \approach\ needs no prior knowledge, not even a cost signal;\approach\ also needs far less data collection than the previously discussed solutions. For example, in \citet{tappler2022automata}, the shielding approach needs approximately $10,000$ episodes to begin promoting safety successfully, whereas we show \approach\ starts safeguarding the RL agent by using only between $500$ to $1,000$ episodes in Figure~\ref{fig:various-episode-results}.


\section{Preliminaries}
\label{sec:prelim}
\textbf{Markov Decision Process.} 
A Markov Decision Process (MDP)~\citep{bellman1957markovian} is a discrete-time stochastic control process to model decision-making. An MDP is formally defined as a 5-tuple $M = (S, A, P, R, \gamma)$, where $S$ is the state space, $s_t$ is the state at timestep $t$, $A$ is the action space, $P$ is the state transition probability matrix/transition function such that $P(s_{t+1}|s_t, a_t)$ is the probability of transitioning to state $s_{t+1}$ from state $s_t$ using action $a_t$, $R$ is the reward function such that $R(s,a)$ is the reward for taking action $a$ in state $s$, and $\gamma$ is the discount factor that determines the weight of future rewards. A policy $\pi:  S \to\Delta(A)$ is a distribution over actions given a state. MDPs can be solved using dynamic programming techniques (e.g., value iteration, policy iteration), which require complete knowledge of the MDP's dynamics~\citep{bertsekas2008introduction}. An MDP includes \emph{accepting states}, which are typically associated with the partial or full completion of a task, as well as \emph{non-accepting terminal states}, which signify the end of the process and are often used to represent safety violations. In this work, we consider MDPs with continuous state and action spaces. That is, $S \subseteq \mathbb{R}^n$ and $A \subseteq \mathbb{R}^m$ are real and possibly high-dimensional values.

\textbf{Reinforcement Learning.} 
Reinforcement Learning (RL) involves training an agent to make a sequence of decisions by interacting with the environment, which is modelled as a Markov Decision Process (MDP). This machine learning technique solves MDPs when full knowledge of the transition function is not available. The agent's goal is to find a policy $\pi^*$ maximising the expected discounted return $E\left[\sum_{t=0}^{\infty}\gamma^{t}R(s_{t}, a_{t})\right]$~\citep{sutton2018reinforcement}. A value function $V_{\pi}(s)$ informs the agent how \textit{valuable} a given state is when following the current policy $\pi$.
Regret, a fundamental concept in RL~\citep{berry1985bandit}, measures the difference between the cumulative reward of an optimal policy and the cumulative reward of the policy followed by the agent, highlighting the cost of learning and exploration over time. The standard regret factor of the RL process, which reflects the optimal trade-off between exploration and exploitation, is $O(\frac{1}{\sqrt{E_\mathit{max}}})$, where $E_\mathit{max}$ is the maximum number of episodes the agent is trained for. Common RL algorithms include Q-learning~\citep{qlearning}, and SARSA~\citep{sarsa}. Deep reinforcement learning (DRL) extends traditional algorithms by utilising deep neural networks to approximate the policy $\pi$ or the value function $V$ when the state/action space is high-dimensional and complex. Actor-critic approaches~\citep{sutton2018reinforcement} are a popular class of algorithms both in traditional and deep RL. Distinctly, the policy (actor) and the value function (critic) are modelled as separate components, allowing for simultaneous updates to both functions. Deep Deterministic Policy Gradient (DDPG)~\citep{ddpg-paper} is an example of an actor-critic approach tailored specifically for continuous action spaces.

\textbf{Contrastive Learning.} 
Contrastive Learning (CL)~\citep{hadsell2006dimensionality} is an unsupervised or semi-supervised machine learning paradigm aimed at distinguishing between similar (positive) and dissimilar (negative) pairs of data points. At its core lies a contrastive loss function, which encourages the model to put similar pairs closer together in the embedding space while separating dissimilar pairs. Given a pair of inputs $x_i$, $x_j$, the contrastive loss function is defined as:
\begin{equation}
\begin{split}
L(x_{i}, x_{j}, y, \theta) &= y \cdot \lVert h_\theta(x_i) - h_\theta(x_j) \rVert^2 + (1-y) \\
&\cdot \max\big(0, m - \lVert h_\theta(x_i) - h_\theta(x_j) \rVert^2\big)
\end{split}
\label{eq:contrastive-loss}
\end{equation}
where the binary label $y$ signifies a similar ($y=1$) or dissimilar ($y=0$) pair, $h_\theta$ is the embedding vector provided by the neural network with weights $\theta$ and the margin~$m$ regulates the minimum distance between dissimilar pairs. The loss function encourages the model to learn meaningful representations that reflect the inherent similarities and differences between data points, thus facilitating the formation of well-defined clusters in the embedding space.


\section{ADVICE}
\label{sec:advice}

The \approach\ post-shielding approach~\footnote{Code available at: https://github.com/team-daniel/ADVICE} (Figure~\ref{fig:advice-overview-large}) empowers safe RL exploration in black-box environments without requiring a system model, domain knowledge or handcrafted rules as in related research~\cite{konighofer2023online,yang2023safe}. The core of \approach\ is a contrastive autoencoder (CA) model that can efficiently distinguish between safe and unsafe features from the feature space $F: S \times A$ (representing all possible state-action pairs). A feature $f_t = (s_t, a_t)$,  $f_t \in F$, denotes the state-action pair at timestep $t$. The CA model leverages a unique loss function where similar and dissimilar features are compared, enabling the systematic identification of meaningful latent feature representations. \approach\ employs these latent representations and trains an unsupervised nearest neighbours model in the learnt embedding space, thus enabling the classification of new features. Formally, the devised \approach\ shield is defined by a function $\phi: F \times \mathbb{Z}_+ \rightarrow A$, such that at time step $t$ during the execution of an episode, \approach\ evaluates the agent's desired action $a_t$ by computing $\phi(f_t, K)$, where $K$ is a parameter that controls the risk aversion level of the shield. This function returns $a_t$ if the desired action is deemed safe (thus endorsing its execution), or returns (and enforces the execution of) another, safe action~$a'_t$ instead. \approach\ considers the agent's performance and automatically adapts the value of $K$, thus supporting the dynamic calibration of \approach's cautiousness level during learning. We detail next the two \approach\ stages, and their use within the RL loop as a post-shield.

\subsection{ADVICE Shield Construction}
\label{sec:advice-training}
As a black-box post-shield, 
\approach\ does not rely on any domain knowledge about the RL agent or its environment. Instead, the shield construction stage is founded on collecting a feature set $F_E$ during an initial, unshielded interaction period of $E$ episodes, where the RL agent is allowed to interact with and collect experience from its environment. This relaxes the safety definition from assuming a cost-based MDP framework (like the approaches from~\cite{csc-agents, ddpglag, tessler2018reward}) to the standard MDP (as in \cite{alshiekh2018safe, belardinelli2025probabilistic}), thereby enabling greater generalisability. A feature $f_t \in F_E$ collected up to episode $E$ is then classified as safe, 
unsafe 
or inconclusive 
based on the following:
\begin{equation}
\label{eq:safe-unsafe-def}
g(f_t)\!\!=\!\! 
\begin{cases} 
\text{safe}, &\hspace{-3mm}\text{if } t = 0 \text{ or } s_{t+1} \text{ is an accepting state}\\
\text{unsafe}, &\hspace{-3mm}\text{if } s_{t+1} \text{ is a non-accepting terminal state}\\
\text{inconclusive}, &\hspace{-3mm}\text{otherwise}
\end{cases}
\end{equation}
%
and the set $F_E$ is partitioned into disjoint subsets of safe, unsafe and inconclusive features:
$F_{safe} = \{f_t \!\in\! F_E | g(f_t) = \text{safe}\}$, $F_{unsafe}\! =\! \{f_t \!\in\! F_E | g(f_t) = \text{unsafe}\}$, 
and $F_{inc}\! =\! F_E \setminus (F_{safe} \cup F_{unsafe})$. 
This partitioning facilitates the contrastive learning of: (i) similar feature pairs (i.e., two safe or two unsafe features); and (ii) dissimilar feature pairs (i.e., a safe and an unsafe feature), allowing the CA model to discern between safe and unsafe features effectively.
This key \approach\ characteristic allows to focus on finding meaningful representations in a lower-dimensional latent space that adequately reflects the problem domain similarities and differences. We assemble a training dataset $\mathcal{C}$ for the CA model by automatically labelling each combination of distinct features from $F_{safe} \cup F_{unsafe}$ as similar (label `1') or dissimilar (`0'): 
\begin{equation}
\begin{aligned}
\mathcal{C} =
   & \{ ((f_t, f_{t'}), 1) \mid  
               f_t, f_{t'}\in F_{safe}\cup F_{unsafe} \wedge \\
               & \qquad\qquad\qquad\qquad g(f_t) = g(f_{t'}) \wedge f_t \neq f_{t'} \}\; \cup\\ 
               & \{ ((f_t, f_{t'}), 0) \mid f_t, f_{t'}\in F_{safe}\cup F_{unsafe} \wedge g(f_t) \neq g(f_{t'})
         \}
\end{aligned}
\label{eq:pair-generation}
\end{equation}
%


Using this dataset, the CA model is trained to optimise two loss functions: contrastive loss (CL) from Equation~\eqref{eq:contrastive-loss} and  mean squared error (MSE). The MSE measures how accurately the model reconstructs input features, while the CL refines the latent space by clustering similar features and separating dissimilar ones based on Euclidean distance. This encourages cohesion among similar features and separation among dissimilar ones. 
Consequently, the \approach\ CA model learns to encode and reconstruct the salient features of the RL problem -- environment state and chosen action -- accurately in a lower-dimensional latent space.

Once trained, the CA model is adept at finding nuanced distinctions between safe and unsafe features, and accurately placing unseen features within the appropriate partitions in the latent space. The shield construction stage of \approach\ concludes with embedding an unsupervised nearest neighbours (KNN) model in the latent space; this model classifies new encoded features as safe/unsafe. A visual representation of the latent data encoding, illustrating the clear separation in the latent (embedding) space, is shown in Figure~\ref{fig:advice-overview-large}.

\begin{algorithm}[tb!]
\caption{ADVICE Execution and Adaptation}
\label{alg:advice-inference}
\begin{algorithmic}[1]
  \REQUIRE Contrastive autoencoder $\mathit{CA}$, Max neighbours $K_{max}$, Safety threshold $K$, Safety violation indicators $(z_i)_{i=1}^E$, Recent/distant time windows $h_r$, $h_d$, Shield construction episodes $E$, Max episodes $E_{max}$, Max timesteps $T$, Policy $\pi$
  \FOR {$e = E+1, E+2,\ldots, E_{max}$} \label{algo:maxE}
    \STATE $s_1 \gets \textsc{Observe()}$, $t\gets 1$
    \WHILE {$t \leq T \wedge s_t \text{ not terminal}$} \label{algo:episodeExec}
      \STATE $a_t \gets \pi(s_t)$ \label{algo:actionExec}
      \begin{tcolorbox}[size=minimal, colback=myorange!10, colframe=myorange, boxrule=0.8pt, left=-2mm, right=-2mm, top=0pt, bottom=0pt, boxsep=5pt, before skip=0pt, after skip=0pt]
      \STATE 
      $f_t \leftarrow (s_t, a_t)$  \label{algo:startShield} 
      \IF{!\textsc{IsSafe}($f_t, K, K_{max},\mathit{CA}$)} \label{algo:unsafeAction}
        \STATE $A_{f_t} \leftarrow 
        \left\{ \bigtimes_{d \in D} A^d_{f_t} \;|\; \forall d \in D \,.\, (A^d_{f_t} \subset A^d) \right\}$ \label{algo:candidateActions}
        \STATE $A_{f_t}' \leftarrow \{a \in A_{f_t} \mid
\textsc{IsSafe}((s_t,a), K, K_{max}, \mathit{CA})\}$\label{algo:candidateFilter}
        \IF{$A_{f_t}' \neq \emptyset$}
          \STATE $a_t \gets \textrm{argmax}_{a \in A_{f_t}'}\; Q_\pi(s_t, a)$ \label{algo:findBestAction}
        \ENDIF
      \ENDIF  \label{algo:endShield}
      \end{tcolorbox}
      \STATE $r_t, s_{t+1} \gets \textsc{Execute}(a_t)$
      \STATE $\pi \gets \textsc{UpdatePolicy}(a_t, s_t, r_t, s_{t+1})$
      \STATE $t\gets t+1$
    \ENDWHILE \label{algo:endFor}
    \begin{tcolorbox}[size=minimal, colback=myorange!10, colframe=myorange, boxrule=0.8pt, left=-2mm, right=-2mm, top=0pt, bottom=0pt, boxsep=5pt, before skip=0pt, after skip=0pt]
    \IF{$s_t \textrm{ is non-accepting terminal}$}
       \STATE $z_e \gets 1$
    \ELSE
       \STATE $z_e \gets 0$
    \ENDIF
    \STATE $K \gets \textsc{UpdateK}(K, K_{max}, (z_i)_{i=1}^e, h_d, h_r)$ \label{algo:updatedCautiousness}
    \end{tcolorbox}
  \ENDFOR \label{algo:endWhile}
  \STATE \;
  \begin{tcolorbox}[size=minimal, colback=myorange!10, colframe=myorange, boxrule=0.8pt, left=-2mm, right=-2mm, top=0pt, bottom=0pt, boxsep=5pt, before skip=0pt, after skip=0pt]
  \STATE \textbf{function} \textsc{IsSafe}($f, K, K_{max}, \mathit{CA}$) )\label{algo:safeStart}
  \STATE \quad $\hat{f} \gets \textsc{Encode}(f, \mathit{CA})$ \label{algo:encoding}
  \STATE \quad $N_{\hat{f}} \gets \textsc{GetSafeNeighbours}(\hat{f}, K_{max})$ \label{algo:safePoints}
  \quad\quad\RETURN $|N_{\hat{f}}|\geq K$\COMMENT{Action safe (unsafe otherwise)}
  \label{algo:safeEnd}
  \end{tcolorbox}
\end{algorithmic}
\end{algorithm}

\subsection{ADVICE Execution and Adaptation}
\label{sec:advice-inference}
Upon completing the shield construction phase, ADVICE, similar to~\cite{tabular-shield}, can be used as a post-shield within the standard RL loop. The constructed shield encourages safe environment exploration, guarding the agent throughout its interaction with the environment by ensuring the execution of safe actions.

The \approach\ execution and adaptation stage (Algorithm~\ref{alg:advice-inference}, the standard RL loop with the \fcolorbox{myorange}{myorange!20}{ADVICE extension}) involves performing the action evaluation and decision-making cycle from lines~\ref{algo:maxE}--\ref{algo:endWhile} until the maximum number of training episodes $E_{max}$ is reached in line~\ref{algo:maxE}. 
During each step $t$ of an episode's execution (lines~\ref{algo:episodeExec}--\ref{algo:endFor}), the RL agent, having observed the current state $s_t$, selects an action $a_t$ based on its current policy $\pi$ (line~\ref{algo:actionExec}).
Without the ADVICE shield, the agent would proceed with this action regardless of its potential safety implications. 
Instead, the ADVICE shield is applied in lines~\ref{algo:startShield}--\ref{algo:endShield} to assess whether the agent-selected action is safe -- and to replace it with a safe action if not, thus ensuring safe environment exploration. This assessment is performed by invoking the function \textsc{IsSafe} in line~\ref{algo:unsafeAction}. 
To establish whether the action associated with a feature $f$ is safe, this function (lines~\ref{algo:safeStart}--\ref{algo:safeEnd}) uses the contrastive autoencoder  $\mathit{CA}$ to extract the latent representation $\hat{f}$ of that feature (line~\ref{algo:encoding}), invokes the function \textsc{GetSafeNeighbours} to collect -- from the $K_{max}$ latent data points nearest to $\hat{f}$ -- those deemed safe into a set $N_{\hat{f}}$ of safe neighbours (line~\ref{algo:safePoints}). Finally, the assessed action is deemed safe if the cardinality of the set $N_{\hat{f}}$ is at least as high as the safety threshold $K$, and unsafe otherwise (line~\ref{algo:safeEnd}).

The safety threshold $K \in [0, K_{max}]$ determines how many neighbours need to be labelled safe for the encoded feature to be deemed safe. 
Values of $K$ close to $K_\mathit{max}$ make \approach\ more risk-averse, while $K$ values closer to $\lceil K_\mathit{max}/2 \rceil$ yield ADVICE shields that are more relaxed and favour exploration. 
A value of $K < \lceil K_\mathit{max}/2 \rceil$ should not be considered, as this would allow the RL agent to execute actions that are more likely to be unsafe than safe.

If the agent-selected action $a_t$ is deemed unsafe (line~\ref{algo:unsafeAction}), \approach\ intervenes and generates a set of valid candidate actions $A_{f_t}$ 
by quantising
each action dimension $A^d$, where $d \in D$ is the dimension, and extracting the Cartesian product across the $D$ action dimensions (line~\ref{algo:candidateActions}).
The candidate actions set $A_{f_t}$ is then filtered into the subset $A'_{f_t} \subseteq A_{f_t}$, to retain only valid and safe actions (line~\ref{algo:candidateFilter}).
If the $A'_{f_t}$ set is not empty, these filtered candidate actions are evaluated by the RL agent's value function $Q_{\pi}$ for their expected reward (line~\ref{algo:findBestAction}), and the action with the highest expected reward is selected as a replacement for the unsafe action $a_t$. We note that this replacement action aligns with both the safety considerations and the agent's performance objectives. 
Finally, if no safe alternative action is identified (i.e., $A_{f_t}'\!=\!\emptyset$), \approach\ retains the originally selected action $a_t$: with no alternative action predicted to be safe, keeping the highest expected reward action $a_t$ is the best option. 

At the end of each episode, \approach\ updates its cautiousness level by considering the recent safety performance of the RL agent (line~\ref{algo:updatedCautiousness}). 
This unique \approach\ characteristic enables moving beyond the static definition of the safety threshold $K$ by automatically adjusting its value based on its recent safe/unsafe behaviour.
Accordingly, \approach\ becomes adaptive and allows the agent to explore more when exhibiting safe behaviour for a period of time while being more cautious, thus interfering more when the RL agent behaves increasingly unsafely.

To assess the agent's performance over time, \approach\ employs a double sliding window, commonly used in the field of anomaly detection
By comparing the current trend of safety violations against a broader historical view, \approach\ can discern whether the current trend deviates from the pattern seen historically. This analysis informs ADVICE to strengthen or relax its cautiousness level during the execution of an episode. The adaptation function is given by:
\begin{equation}
	K =
	\begin{cases} 
		\min(K\!+\!1, K_{max}),    &\text{if } \mu_{r}\!\!>\!\!\mu_{d}\!+\!\sigma_{d} \\
		\max(K\!-\!1, \lceil K_{max}/2\rceil ), & \text{if } \mu_{r}\!\!<\!\!\mu_{d}-\sigma_{d} \\
		K, & \text{otherwise}
	\end{cases}
	\label{eq:updatek}
\end{equation}
where 
\begin{equation}
   \mu_{r}\!=\!\frac{\sum_{i=e-h_{r}+1}^{e}z_i}{h_r},
   \mu_{d}\!=\!\frac{\sum_{i=e-h_{d}+1}^{e}z_i}{h_d},
  \sigma_{d}\!=\!\sqrt{\mu_{d}(1-\mu_{d})}
\label{eq:mov-metrics}
\end{equation}
represent the moving probabilities of an episode ending with a safety violation $z_e$ in the recent $h_r$ and distant $h_d$ past, and the standard deviation for the former, respectively.

If the $\mu_{r}$ value exceeds the sum of $\mu_{d} + \sigma_{d}$, this signifies that the agent incurs safety violations more often than expected. Consequently, $K$ is automatically incremented to adopt a more cautious position. Conversely, if $\mu_{r}$ is less than $\mu_{d} - \sigma_{d}$, it is an indication that the agent acts more cautiously than expected. As a result, $K$ is automatically reduced to allow the agent to explore more freely. The safety threshold $K$ remains the same for any other scenario. 
\approach\ deliberately only considers one standard deviation as two or more standard deviations would make the adaptation slow to respond to emerging safety risks. Our sensitivity analysis on these two parameters (Appendix~\ref{sec:appendix-adaptive}), corroborates the described behaviours.


\section{Theoretical Analysis}
Through theoretical analysis, we show that the misclassification probability of unseen features is bounded and decreases with more diverse data collected up to episode~$E$. We also derive regret bounds for \approach\ with fixed and adaptive~$K$, showing that adaptive~$K$ yields lower long-term regret. All theorem proofs are in Appendix~\ref{sec:appendix-theo-analysis}.

\begin{theorem}
The probability of \approach\ misclassifying a feature is bounded by $exp(-\frac{\gamma^2}{2\sigma^2})$, where $\gamma$ is the contrastive separation margin and $\sigma^2$ is the variance of the 
Gaussian noise within the latent space.
\label{thm:1}
\end{theorem}

\begin{figure*}[tb]
\centering
\begin{minipage}{\textwidth}
    \centering
    \includegraphics[width=\linewidth]{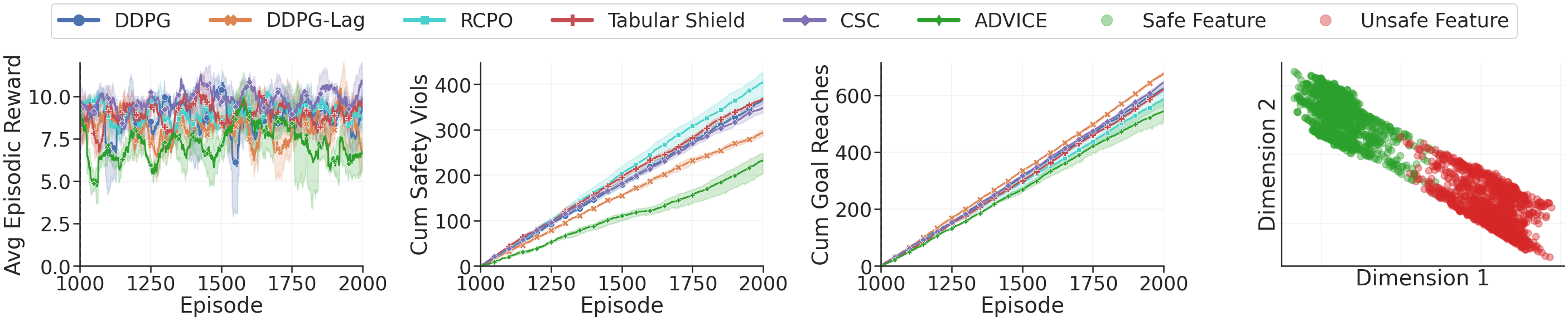}
    \smallskip
    \textbf{(a)} Semi-random goal environment results
\end{minipage}
\begin{minipage}{\textwidth}
    \centering
    \includegraphics[width=\linewidth]{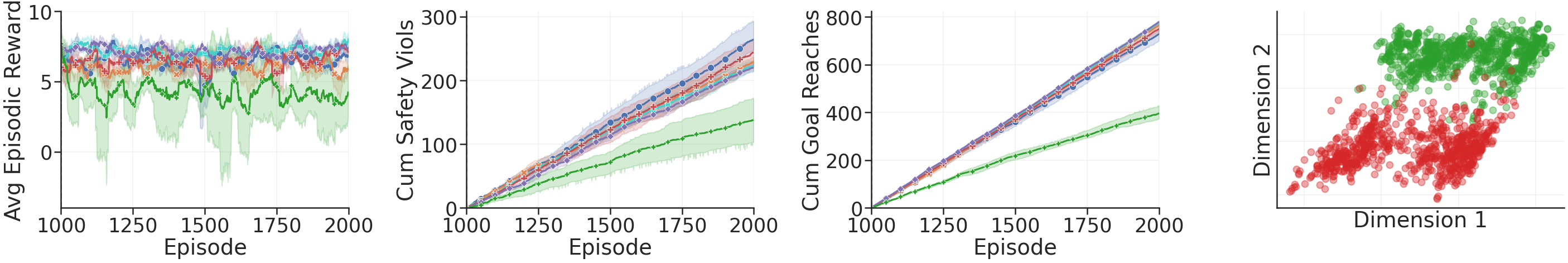}
    \smallskip
    \textbf{(b)} Randomised goal environment results
\end{minipage}
\begin{minipage}{\textwidth}
    \centering
    \includegraphics[width=\linewidth]{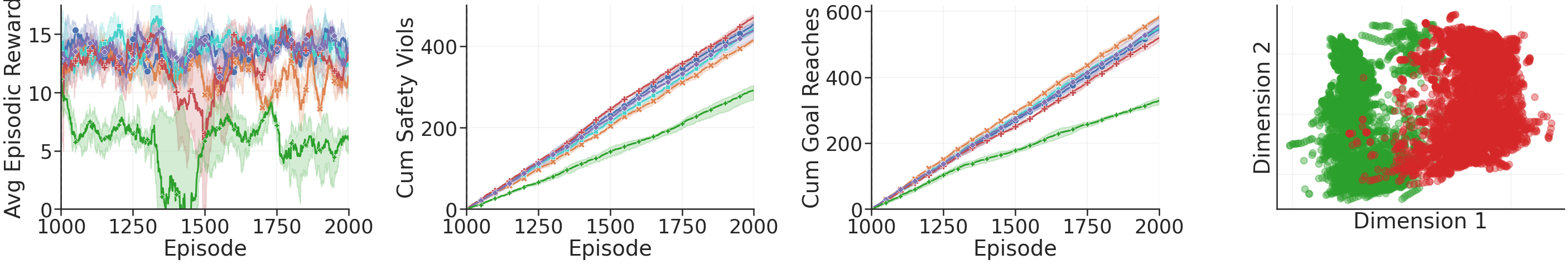}
    \smallskip
    \textbf{(c)} Randomised button environment results
\end{minipage}
\begin{minipage}{\textwidth}
    \centering
    \includegraphics[width=\linewidth]{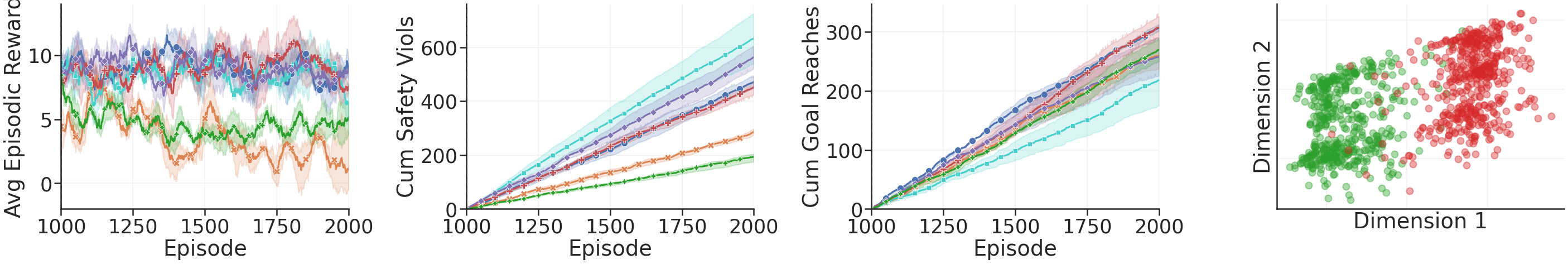}
    \smallskip
    \textbf{(d)} Randomised circle environment results
\end{minipage}
\caption{Average episodic reward, cumulative safety violations, cumulative goal reaches of examined approaches (DDPG, DDPG-Lag, RCPO, Tabular shield, CSC, ADVICE) and example latent visualisation for the semi-random goal (a), randomised goal (b), randomised button (c), and randomised circle (d) environments. Results are shown for episodes $1000$–$2000$ to highlight ADVICE's execution and adaptation (Figure~\ref{fig:advice-overview-large}-B), with full results shown in Appendix~\ref{sec:appendix-full-results} (Figure~\ref{fig:appendix-full-results}).}
\label{fig:core-results}
\end{figure*}

Theorem~\ref{thm:1} shows that ADVICE's misclassification probability decreases exponentially with a larger contrastive separation margin $\gamma$ and smaller noise $\sigma^2$.
This emphasizes the importance of training an efficient encoder 
during \approach's shield construction stage.

\begin{theorem}
The probability of \approach\ misclassifying a feature decreases exponentially with improved data entropy and is bounded by $exp(-\frac{H(F_E)}{2\sigma^2})$.
\label{thm:2}
\end{theorem}
Theorem~\ref{thm:2} extends Theorem~\ref{thm:1} by highlighting that diverse and representative data collected up to episode $E$ reduces misclassification rates, with dependence on the entropy $\sqrt{H(F_E)}$ of the feature set.

\begin{theorem}
The expected regret of \approach\ with a fixed $K$ is upper bounded by a function of order $O\left(\frac{1}{\sqrt{E_{max}}}\right) + O\left(E_{max} \cdot \Delta Q \cdot exp\left(- \frac{\gamma^2}{2\sigma^2} \right) \cdot \frac{1}{\sqrt{K}} \right)$.
\label{thm:3}
\end{theorem}

Theorem~\ref{thm:3} establishes the baseline regret for ADVICE with a fixed $K$. The dependence on $K$ in the second term $\frac{1}{\sqrt{K}}$ highlights the trade-off between a cautious shield (larger $K$) and a relaxed one (smaller $K$).

\begin{theorem}
The expected long-term regret of \approach\ with an adaptive $K$ is lower than that of \approach\ with a fixed $K$, and is upper bounded by a function of order $O\left(\frac{1}{\sqrt{E_{max}}}\right) + O \left(E_{max} \cdot \Delta Q \cdot exp\left(- \frac{\gamma^2}{2\sigma^2} \right) \cdot \frac{1}{\sqrt{K^*}} \right) + O \left(\frac{1}{E_{max}} \sum^{E_{max}}_{e=1} |K_e - K^* | \cdot \Delta Q_{K_{e}} \right)$.
\label{thm:4}
\end{theorem}

Theorem~\ref{thm:4} shows that adaptive $K$ achieves lower long-term regret than fixed $K$. The penalty term $|K_e - K^*|$ captures the transient cost of tuning $K$ dynamically; however, this diminishes over time as $K \rightarrow K^*$ approaches the optimal $K^*$. Adaptive $K$ effectively balances safety and exploration, improving long-term.


\begin{figure*}[tb!]
\centering
\begin{minipage}{0.15\textwidth}
    \centering
    \includegraphics[width=\textwidth]{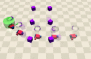}
    
    \smallskip
    \textbf{(a)} DDPG
    \label{fig:ddpg-traj}
\end{minipage}
\hfill
\begin{minipage}{0.15\textwidth}
    \centering
    \includegraphics[width=\textwidth]{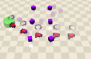}
    
    \smallskip
    \textbf{(b)} DDPG-Lag
    \label{fig:ddpg-lag-traj}
\end{minipage}
\hfill
\begin{minipage}{0.15\textwidth}
    \centering
    \includegraphics[width=\textwidth]{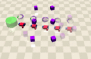}
    
    \smallskip
    \textbf{(c)} RCPO
    \label{fig:rcpo-traj}
\end{minipage}
\hfill
\begin{minipage}{0.15\textwidth}
    \centering
    \includegraphics[width=\textwidth]{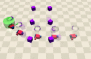}
    
    \smallskip
    \textbf{(d)} Tabular Shield
    \label{fig:tab-shield-traj}
\end{minipage}
\hfill
\begin{minipage}{0.15\textwidth}
    \centering
    \includegraphics[width=\textwidth]{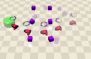}
    
    \smallskip
    \textbf{(e)} CSC
    \label{fig:csc-traj}
\end{minipage}
\hfill
\begin{minipage}{0.15\textwidth}
    \centering
    \includegraphics[width=\textwidth]{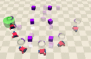}
    
    \smallskip
    \textbf{(f)} ADVICE
    \label{fig:advice-traj}
\end{minipage}
\caption{Example trajectories of the assessed approaches on the semi-random goal environment; obstacles are \textcolor{violet}{purple}, goals are \textcolor{green}{green} circles, and the RL agent is the \textcolor{red}{red} vehicle.}
\label{fig:example-trajectories}
\end{figure*}

\begin{figure*}[tb]
    \centering
    \includegraphics[width=0.95\linewidth]{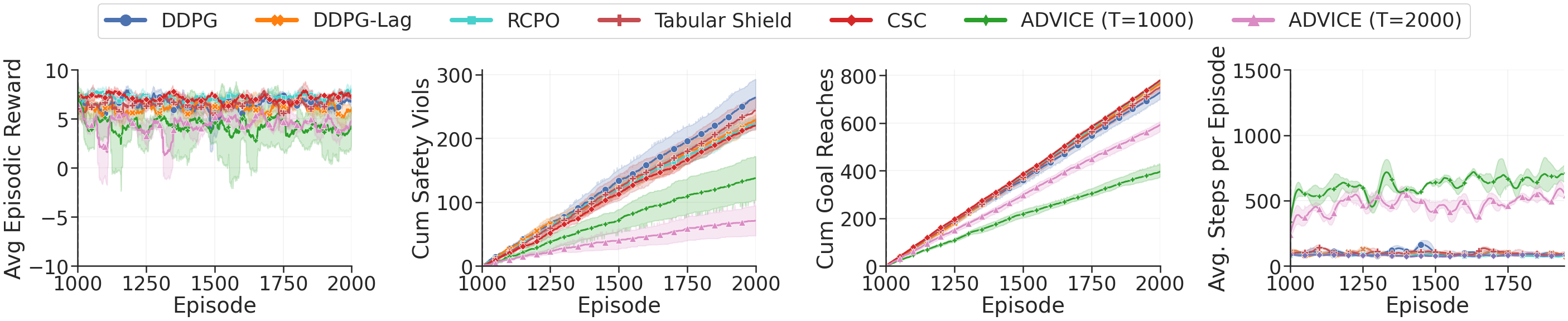}
    \caption{Average episodic reward, cumulative safety violations, and cumulative goal reaches of various approaches on the randomised goal environment where the maximum episodic steps are doubled.}
    \label{fig:various-steps-results}
\end{figure*}

\section{Evaluation}
\label{sec:evaluation}

We evaluate ADVICE using tasks from the Safety Gymnasium test-suite~\citep{safety-gymnasium}, where a robot with Lidar sensors navigates environments with multiple obstacles. This benchmark assesses performance and safety in high-dimensional scenarios with various constraints, enabling a fair comparison of \approach\ against approaches that also used this benchmark~\citep{csc-agents}. A terminal state occurs when the agent collides with an obstacle, causing catastrophic damage and ending the episode. The selected tasks cover varying complexities in agent, goal, and obstacle positions:

\begin{itemize}[noitemsep, nolistsep, leftmargin=3mm]
    \item \textbf{Semi-random Goal}: 
    A goal environment with six static obstacles, with the agent and goal spawning randomly in each episode.
    \item \textbf{Randomised Goal}: 
    Similar to the Semi-random Goal environment, with obstacles also having random positions per episode.
    \item \textbf{Randomised Circle}: The agent has to circle in a given zone, aiming to maximise speed and distance from the zone's centre while avoiding three randomised obstacles.
    \item \textbf{Randomised Button}: The agent must press the correct button in an environment with six obstacles and four buttons, with random positions per episode. Pressing the wrong button ends the episode.
    \item \textbf{Constrained Randomised Goal}: A randomised goal environment with hazards that impose a step-wise cost when the agent is inside. The aim in this environment is to minimise the cost.
\end{itemize}

Details of these environments and tasks are in Appendix~\ref{sec:appendix-task-details}, with Figure~\ref{fig:examples-envs} showing task instances from the Safety Gymnasium. 

To evaluate ADVICE, we compare it with the following state-of-the-art algorithms:
\begin{itemize}[noitemsep, nolistsep, leftmargin=3mm]	
    \item \textbf{DDPG}: A deep deterministic policy gradient agent ~\citep{ddpg-paper} which is  the foundational baseline.
    \item \textbf{DDPG-Lag}: A DDPG agent with an online Lagrangian multiplier ($\lambda\!=\!0.1$, $\alpha\!=\!0.01$) which heavily penalises constant safety violations~\citep{ddpglag} dynamically.
    \item \textbf{RCPO}: A reward-constrained policy optimisation framework, where a Lagrangian balances reward maximisation and constraint satisfaction~\citep{tessler2018reward}.
    \item \textbf{Tabular Shield}: A DDPG agent with a tabular shield storing terminal state-action pairs, preventing actions within 0.1 of recorded pairs~\citep{tabular-shield}.
    \item \textbf{Conservative Safety Critic (CSC)}: A DDPG agent with a conservative safety critic that uses conservative estimates to evaluate safety~\citep{csc-agents}. 
\end{itemize}

Other black-box safe exploration approaches discussed in Section~\ref{sec:rel-work} were excluded because they either represent earlier research from approaches used in our evaluation or due to substantial data collection requirements. 
To enable a fair comparison against the evaluated approaches, they use the same DDPG configuration; algorithm details and hyperparameter information are provided in Appendix~\ref{sec:appendix-hyperparams}. 
Following~\cite{dalal2018safe, csc-agents}, we perform five independent runs, with results showing mean scores and confidence intervals (standard error). Unless stated otherwise, the default deployment \approach\ per run is $E\!=\!1000$, $K=4$, $K_{max}=5$, $T=1000$, $h_d=25$, and $h_r=3$. 

To evaluate safe RL approaches under a common setting, we adopt unconstrained, black-box MDPs, where constraint violations (non-accepting terminal states) are sparse. In this setup, \approach\ collects features upon encountering a non-accepting terminal state, while constrained approaches (e.g., DDPG-Lag, RCPO, CSC) receive explicit cost signals at each violation. Upon encountering a non-accepting terminal state, the episode terminates and is reset.
To further ensure fairness, we also evaluate all approaches in a fully constrained MDP environment (Constrained Randomised Goal, see Figure~\ref{fig:appendix-cmdp-results}), where explicit costs are assigned during an episode. In this setting, \approach\ collects features upon observing a cost, conforming to the assumptions of constrained approaches during training.

\subsection{Performance Results}
\label{sec:core-results}

\begin{figure*}[t]
    \centering
    \includegraphics[width=0.95\linewidth]{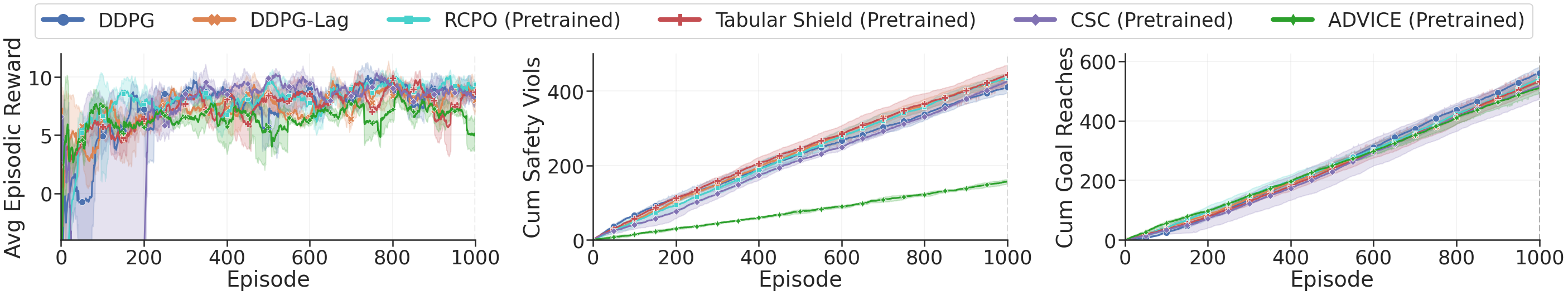}
    \caption{Average episodic reward, cumulative safety violations, and cumulative goal reaches on the semi-randomised goal environment where \approach, Tabular Shield, and CSC have been pre-trained on the randomised goal environment to show transfer learning and distributional shift robustness capabilities.}
    \label{fig:appendix-transfer-learning}
\end{figure*}

\begin{figure*}[h!]
    \centering
    \includegraphics[width=0.95\linewidth]{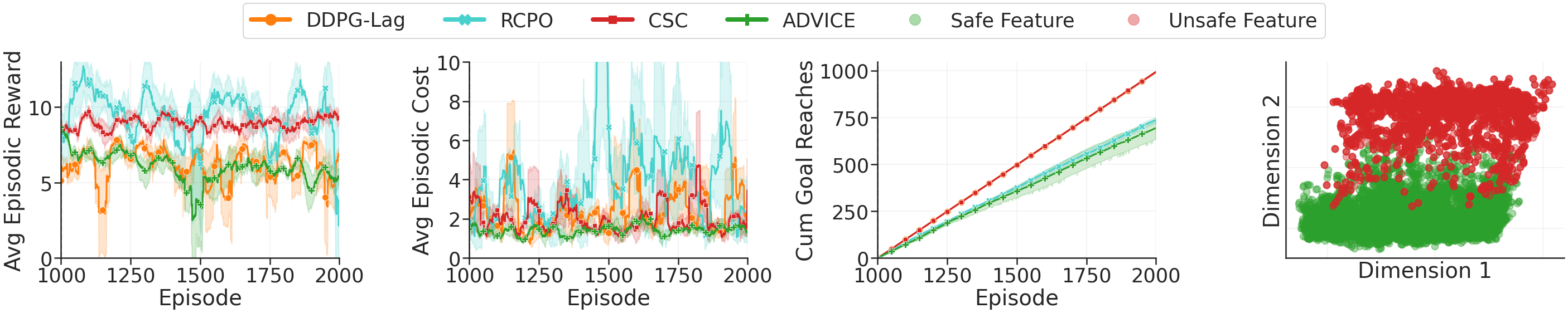}
    \caption{Average episodic reward, average episodic cost, cumulative goal reaches, and example latent space visualisation on the constrained randomised goal environment.}
    \label{fig:cmdp-results}
\end{figure*}

For our experiments, we evaluate the performance of the comparable approaches in all environments. Figure~\ref{fig:core-results} shows average episodic reward, cumulative safety violations, and cumulative goal reaches; the rightmost figure shows the latent space visualisation from a single ADVICE run. Full results are in Appendix~\ref{sec:appendix-full-results}. The results show that \approach\ prioritises safety and achieves competitive rewards across tasks. Both DDPG-Lag and \approach\ exhibit a trade-off between reward maximisation and safety.

The primary goal of a safety shield is to enable safe exploration and learning. Across all tasks, \approach\ significantly reduces safety violations, outperforming other approaches by a notable margin. While DDPG-Lag also reduces violations, it is less effective than \approach, highlighting the strength of \approach's shield. RCPO and CSC reduce safety violations by a small amount but vastly underestimate safety due to sparse data in unconstrained MDPs with complex constraints, where regular neural network-based approaches struggle without a consistent cost signal. This demonstrates the strength of \approach's contrastive learning. The learnt latent space from an \approach\ run (Figure~\ref{fig:core-results}-column 4) further validates its ability to distinguish safe and unsafe features. Despite requiring an initial unshielded exploration phase ($E\!=\!1000$ episodes) to construct its shield, \approach\ ultimately achieves fewer safety violations than DDPG-Lag, which applies its constraints from episode $E\!=\!1$. This highlights the superior long-term effectiveness of \approach's shielding mechanism.

Despite cautious prioritisation, \approach\ achieves similar goal-reaching frequencies as other approaches while maintaining safety. In fact, \approach\ completes the task without compromising safety, and the difference in accumulated reward is due to the reward function design and the reduced time the agent has to complete the task using a cautious path. Through additional experiments with an increased number of maximum steps per episode ($T\!=\!2000$ -- Figure~\ref{fig:various-steps-results}), we have established that \approach's average episodic return is very similar to the other approaches. Hence, increasing the episode's duration yields improved \approach-based results. These results confirm that \approach's contrastive shielding effectively reduces safety violations compared to other approaches, \changed{with minimal impact on the agent's overall performance.}

Figure~\ref{fig:example-trajectories} shows example trajectories in the semi-random goal environment, further 
supporting our findings. DDPG and Tabular shield prioritise return maximisation, navigating through the centre of the object wall, which is risky due to action noise, as seen in Figure~\ref{fig:core-results}a. Cost-based approaches learn similar behaviours but maintain a larger gap from objects to account for noise. 
\changed{In contrast, \approach\ follows a cautious path around the wall, reducing collision risk more than any other method. This longer trajectory lowers the average episodic reward, since the agent initially moves away from the goal to avoid obstacles, and the reward function penalises such detours while neglecting safety risk.}


\subsection{Transfer Learning}
\label{sec:transfer-learn}

\changed{Figure~\ref{fig:appendix-transfer-learning} shows \approach's transfer learning and robustness under distribution shift in a semi-random environment. Using a pre-trained shield from the randomized setting, it cuts safety violations by over 50\% across 1000 episodes, clearly outperforming RCPO, Tabular Shield, and CSC. These results demonstrate \approach's ability to generalise its learned shield to similar, non-stationary environments while maintaining comparable reward and goal-reaching rates, highlighting both robustness and effectiveness.}


\subsection{Constrained Environment}
\label{sec:CMDP}

We also demonstrate \approach's ability to operate in unconstrained MDP environments (albeit being less common than constrained ones). 
Figure~\ref{fig:cmdp-results} shows results in the Constrained Randomised Goal environment (full results in Appendix~\ref{sec:appendix-full-results}).
\approach\ consistently achieved the lowest average episodic cost, outperforming cost-based approaches, which showed significant oscillations, highlighting their difficulty in balancing cost reduction with reward maximization~\citep{liu2022constrained}. Notably, such oscillations were absent in sparse-cost environments like the unconstrained MDP environments discussed in Section~\ref{sec:core-results}.


\section{Conclusion and Future Work}
\label{sec:conc-and-future}
\changed{
\approach\ is a post-shielding approach for the safe exploration of RL agents operating in complex \textit{black-box} environments. 
\approach\ does not need \emph{any} domain knowledge or \emph{any} scenario-specific safety definitions, thus yielding improved generalizability. 
Our evaluation shows that \approach\ significantly reduces safety violations while maintaining competitive performance against state-of-the-art approaches. 
Our future work involves exploring meta-learning techniques 
for faster \approach\ activation without sacrificing performance. 
Quantisation 
could reduce computational demands, enhancing applicability in computationally constrained domains, while using uncertainty quantification~\cite{abdar2021review,bethell2024robust} in state estimation could yield more robust policies.
\approach\ could become a foundational step in using neural network-based shielding for safe RL exploration in complex, black-box environments without \emph{any} domain knowledge.
}



\begin{ack}
This research has received funding from the Doctoral Centre for Safe, Ethical and Secure Computing (SEtS) at the University of York, UK, the European Union’s Horizon projects GuardAI and AI4Work (grant agreements No 101168067 and 101135990, respectively), and the UK Advanced Research and Invention Agency's Safeguarded AI project ULTIMATE. This work was also supported by the Centre for Assuring Autonomy, University of York, UK. 
The authors are grateful to Charmaine Barker for her help and insight on this paper.
\end{ack}


\bibliography{references}


\newpage
\onecolumn
\appendix


\section{Task Details}
\label{sec:appendix-task-details}
\begin{figure*}[h]
\centering
\begin{minipage}{0.65\textwidth}
    \begin{minipage}[t]{0.24\linewidth}
        \centering
        \includegraphics[width=\linewidth]{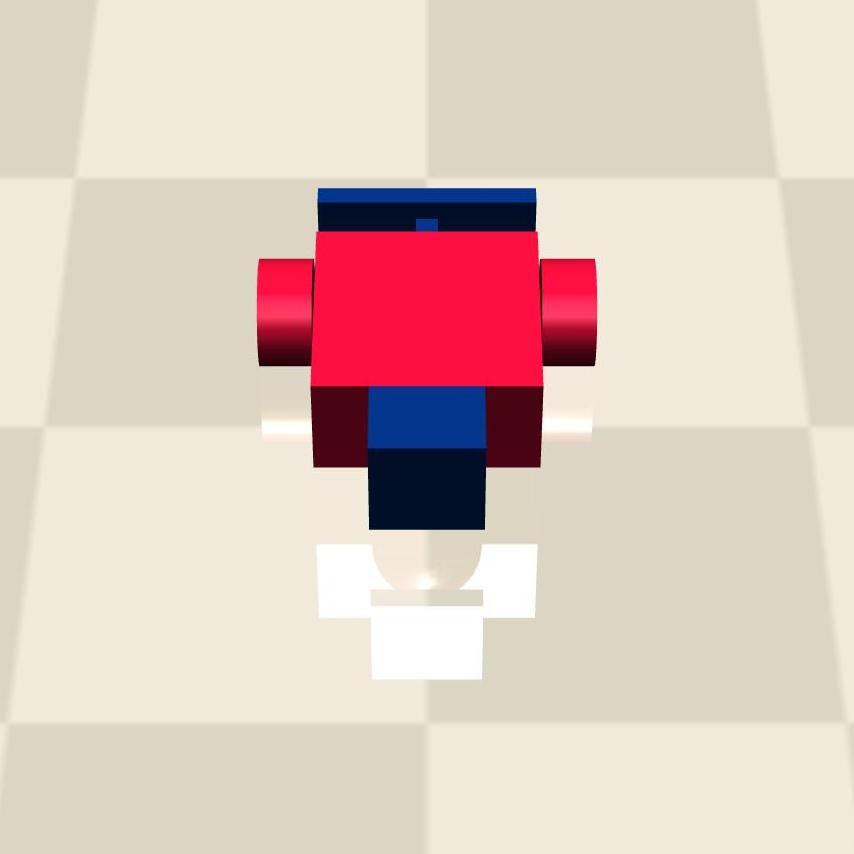}
        \smallskip
        \textbf{(a)} Front
    \end{minipage}
    \hfill
    \begin{minipage}[t]{0.24\linewidth}
        \centering
        \includegraphics[width=\linewidth]{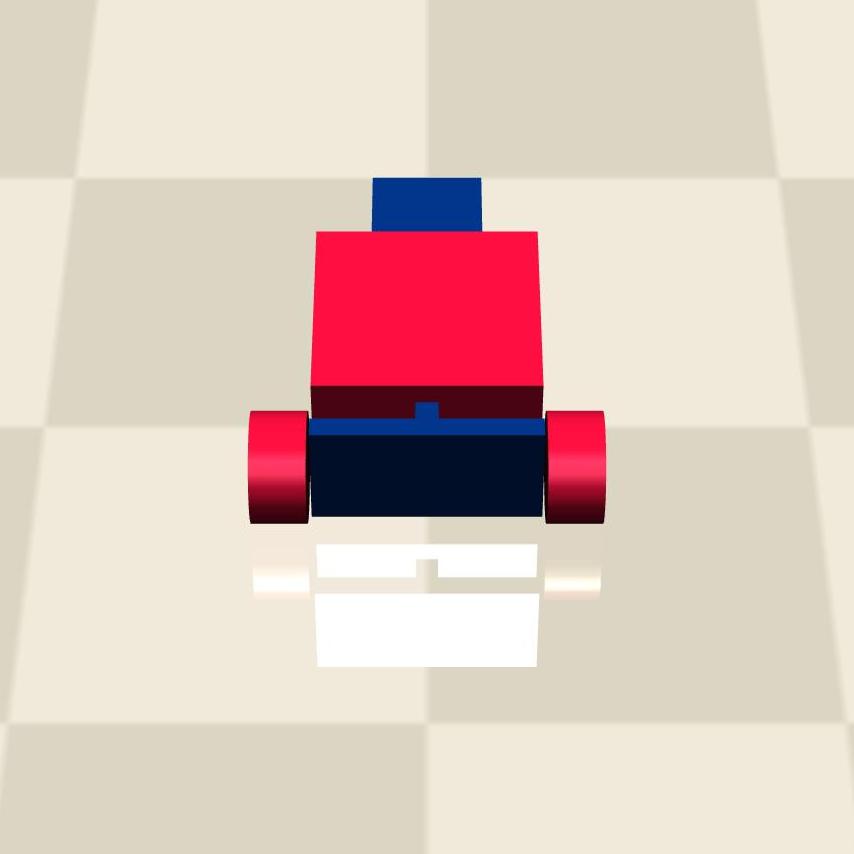}
        \smallskip
        \textbf{(b)} Back
    \end{minipage}
    \hfill
    \begin{minipage}[t]{0.24\linewidth}
        \centering
        \includegraphics[width=\linewidth]{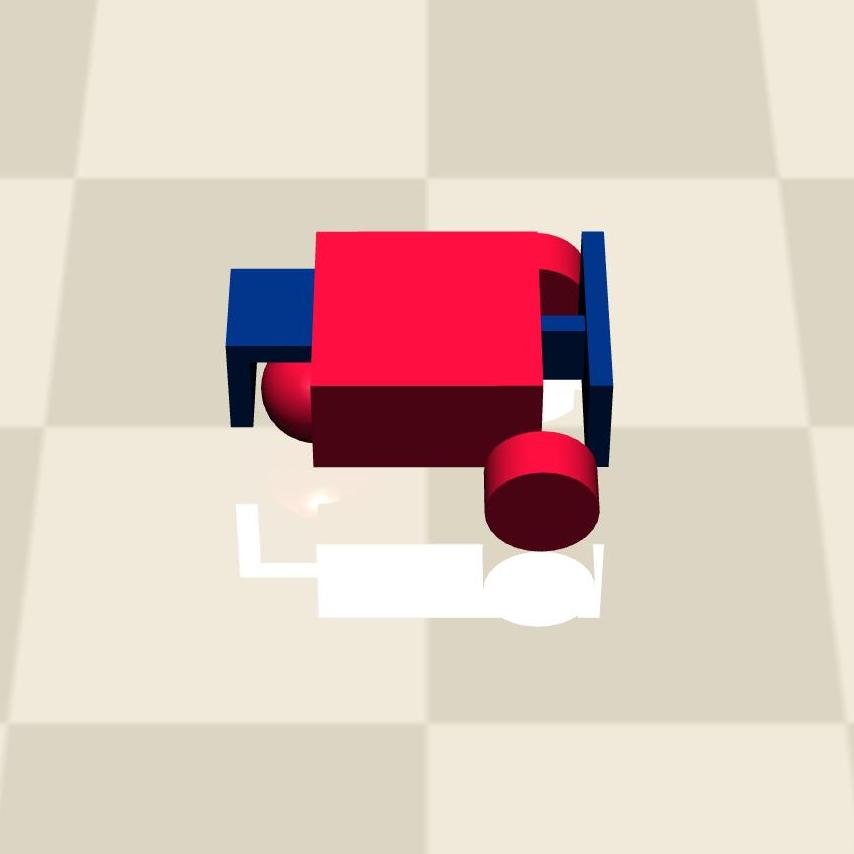}
        \smallskip
        \textbf{(c)} Left
    \end{minipage}
    \hfill
    \begin{minipage}[t]{0.24\linewidth}
        \centering
        \includegraphics[width=\linewidth]{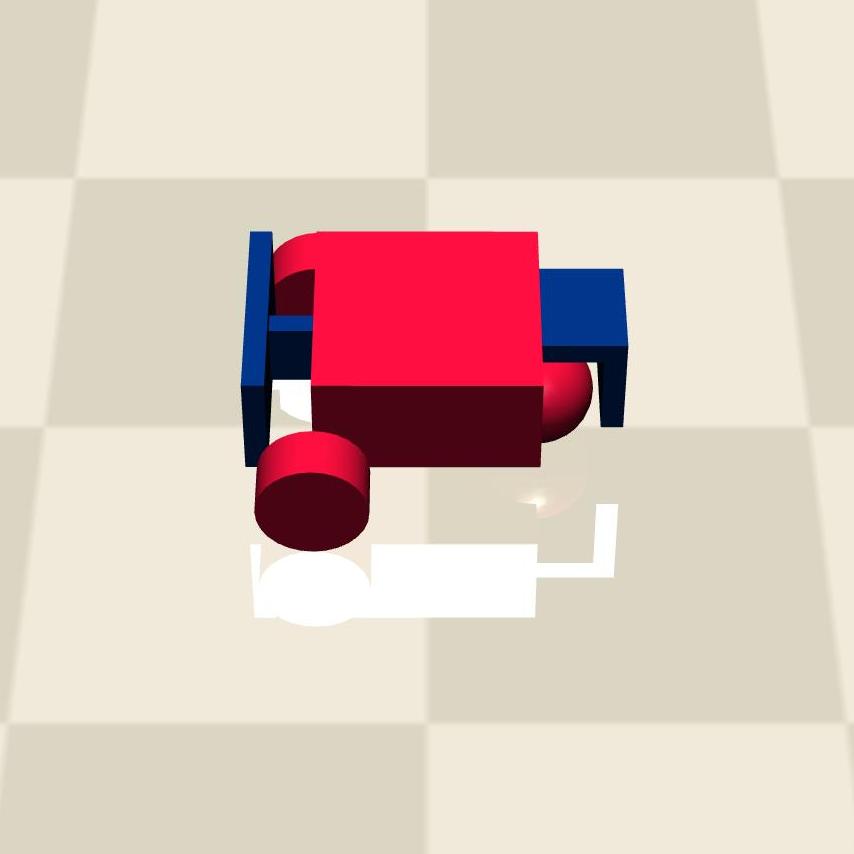}
        \smallskip
        \textbf{(d)} Right
    \end{minipage}
\end{minipage}
\caption{Different views of the Car robot in the Safety Gymnasium test suite~\citep{safety_gym_cars}.}
\label{fig:appendix-car-robot}
\end{figure*}

In the Safety Gymnasium~\citep{safety-gymnasium}~\footnote{More details here: https://github.com/PKU-Alignment/safety-gymnasium} test-suite, a robot with Lidar sensors has to navigate through environments with obstacles to complete a given task. The test suite comes with a multitude of robots (e.g. Point, Ant, Car) and a set of tasks (e.g. Goal, Circle, Button) that can be evaluated. In our experiments, found in Section~\ref{sec:evaluation}, we chose to use:

\begin{itemize}[noitemsep, nolistsep]
    \item \textbf{Car robot:} This robot has two wheels on the rear that the agent can control with one free-rolling front wheel. Steering and movement require nuanced coordination. The action space for the car is $\left[-1, 1\right]^{2}$, and the agent is shown in Figure~\ref{fig:appendix-car-robot}.
    \item \textbf{Semi-random Goal}: A standard goal environment, where the agent aims to reach the goal at the end of the episode whilst navigating through six obstacles. The six obstacles have a static spawn, the agent and the goal have randomised positions every episode. We placed the six obstacles to form a large \textit{wall}, where the agent can fit through to reach the goal but with an increased risk of crashing. In this instance, we want to determine if \approach\ and other safe RL exploration approaches will learn to avoid the \textit{wall} or risk going through it. Deployment trajectories in Figure~\ref{fig:example-trajectories}, show the learnt trajectories of \approach\ and other approaches. 
    \item \textbf{Randomised Goal}: This environment is similar to the semi-random goal environment, with the additional complexity of the obstacles also having random spawns. This extra randomised aspect adds increased difficulty as the agent and safety mechanisms cannot memorise the positions of the obstacles to avoid.
    \item \textbf{Randomised Button}: The environment has two pillar obstacles, four regular obstacles, and four buttons all with random positions and rotations every episode. The agent has to press the correct button which is selected randomly per episode. If the agent collides with an obstacle or presses the incorrect button, the episode will terminate. This scenario increases the task difficulty as the safe and unsafe features collected share significant overlap.
    \item \textbf{Randomised Circle}: The agent has to circle in a given zone in this environment. The aim is to maximise speed and distance from the centre of the zone whilst navigating through three randomised obstacles. This scenario element further increases the task difficulty as now the obstacles to avoid are directly within the area where the agent can maximise its reward. 
    \item \textbf{Constrained Randomised Goal}: This environment is similar to the random goal environment, but instead the task is set up as a constrained MDP. Instead of obstacles that terminate the episode, the task includes hazards that give the agent $0.2$ cost per step when inside them. The agent cannot terminate in this task and instead has to minimise cost whilst maximizing rewards.
\end{itemize}

In all environments for all the tasks above, the agent uses \textit{psuedo} Lidar to perceive objects in the environment. Each type of object (e.g. goal, obstacles) in the environment has its own separate Lidar observation, where a Lidar vector has 16 bins. All vectors are flattened into one observational vector and then given to the agent as the current state. For example, in the semi-random goal environment, there is a goal and a set of obstacles. Here the observational space is $\left[0, 1\right]^{32}$. All lidar vectors have a max distance of $3$ meters. In both goal environments, an episode has a maximum timestep of $1000$. In the circle environment, the maximum timestep is $500$. A \textit{goal reach} in both goal environments is defined as reaching the goal before the episode truncates, in the circle environment it is defined as being within the circle when the episode truncates. A \textit{crash} is defined as the agent colliding with an obstacle, when this occurs, the episode is terminated shortly. Anything else is considered as the episode \textit{timing out}. For the cost-based approaches in the unconstrained environments, the approaches receive a cost of $0.2$ when entering a non-accepting terminal state. This was chosen via a mixture of manual tuning, the recommended values from Safety Gymnasium, and comparative approaches.

Each task has a separate reward function for the agent to maximise. Whenever the agent comes into contact with an obstacle, a constraint cost of $-1$ is given (the exception being the Constrained Randomised Goal task). The reward functions for each task are:

\begin{itemize}[noitemsep, nolistsep]
    \item \textbf{Semi-random, Randomised Goal, Randomised Button, \& Constrained Randomised Goal}: $R_t = (D_{last} - D_{now})\beta$, where $D_{last}$ is the distance between the agent and the goal in timestep $t_{-1}$, $D_{now}$ is the distance between the agent and the goal in timestep $t$, and $\beta$ is the discount factor. Simply, the agent moving towards the goal, in terms of Euclidean distance, gains a positive reward. The agent moving away from the goal gains a negative reward. Reaching the goal gains a static reward of $+1$.
    \item \textbf{Randomised Circle}: $R_t = \frac{1}{1 + |r_{agent} - r_{circle}|} * \frac{(-uy + vx)}{r_{agent}}$ where $(u, v)$ is the x-y velocity coordinates of the agent, $(x, y)$ are the x-y coordinates, $r_{agent}$ is the Euclidean distance of the agent from the origin of the circle, and $r_{circle}$ is the radius of the circle. Simply, the agent is rewarded for moving at speed along the circumference of the circle.
\end{itemize}

It should be emphasized that the agent or approaches used in the evaluation have no prior knowledge of the task/environment/safety concern. Thus, we can define this environment and all tasks within as \textit{black box.}

\begin{figure*}[t]
\centering
\begin{minipage}[t]{0.19\textwidth}
    \centering
    \includegraphics[width=\textwidth]{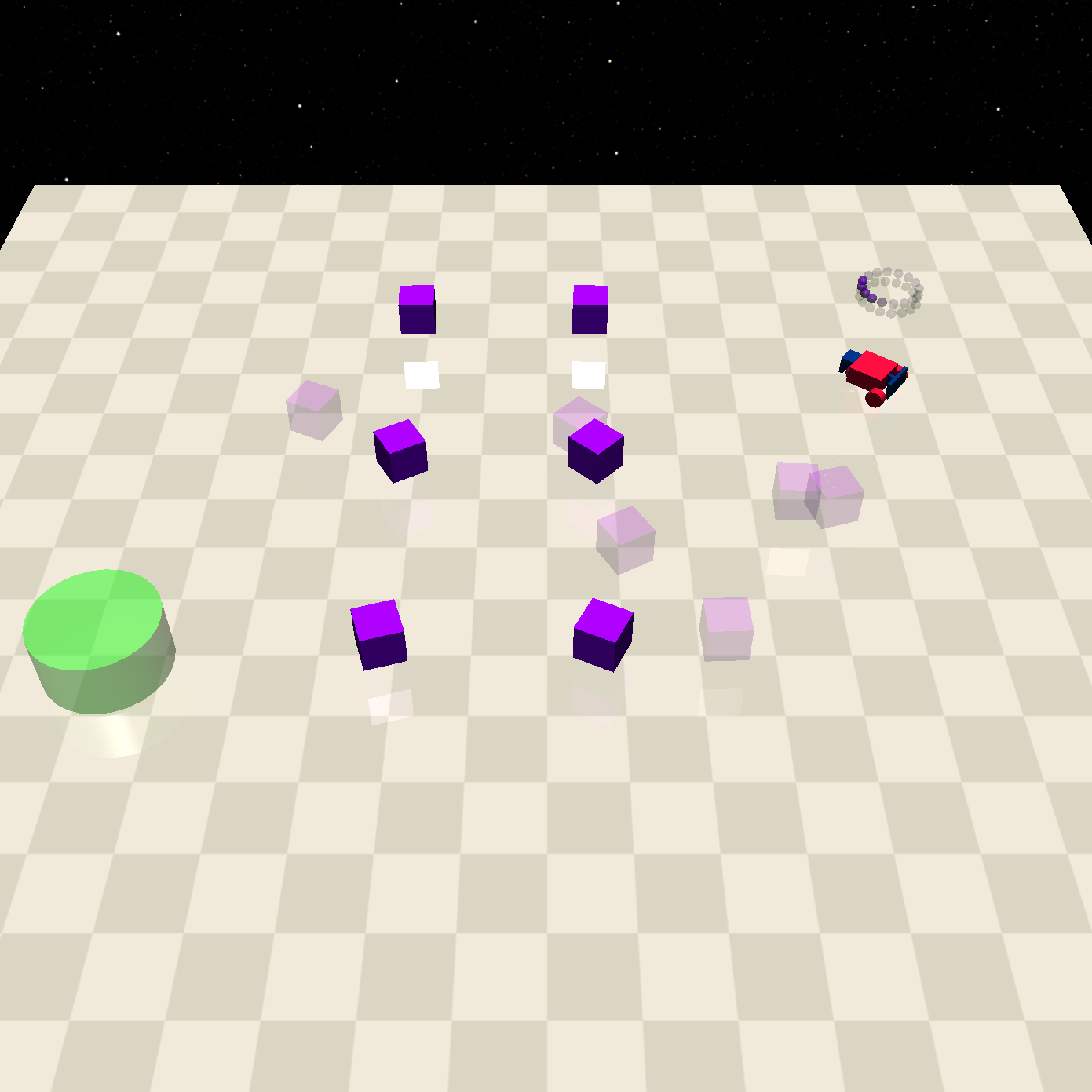}
    
    \smallskip
    \textbf{(a)} Semi-random Goal
\end{minipage}
\hfill
\begin{minipage}[t]{0.19\textwidth}
    \centering
    \includegraphics[width=\textwidth]{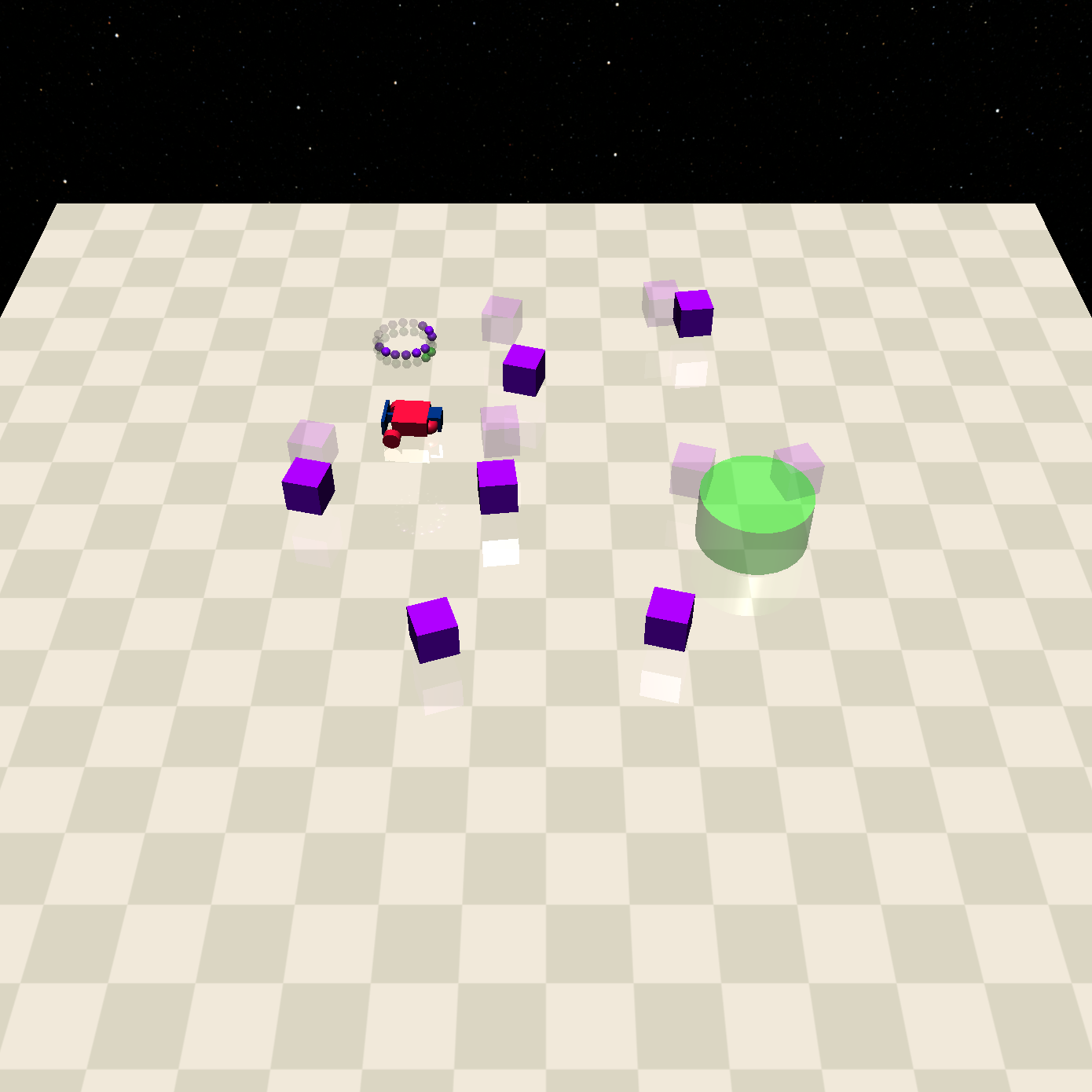}
    
    \smallskip
    \textbf{(b)} Randomised Goal
\end{minipage}
\hfill
\begin{minipage}[t]{0.19\textwidth}
    \centering
    \includegraphics[width=\textwidth]{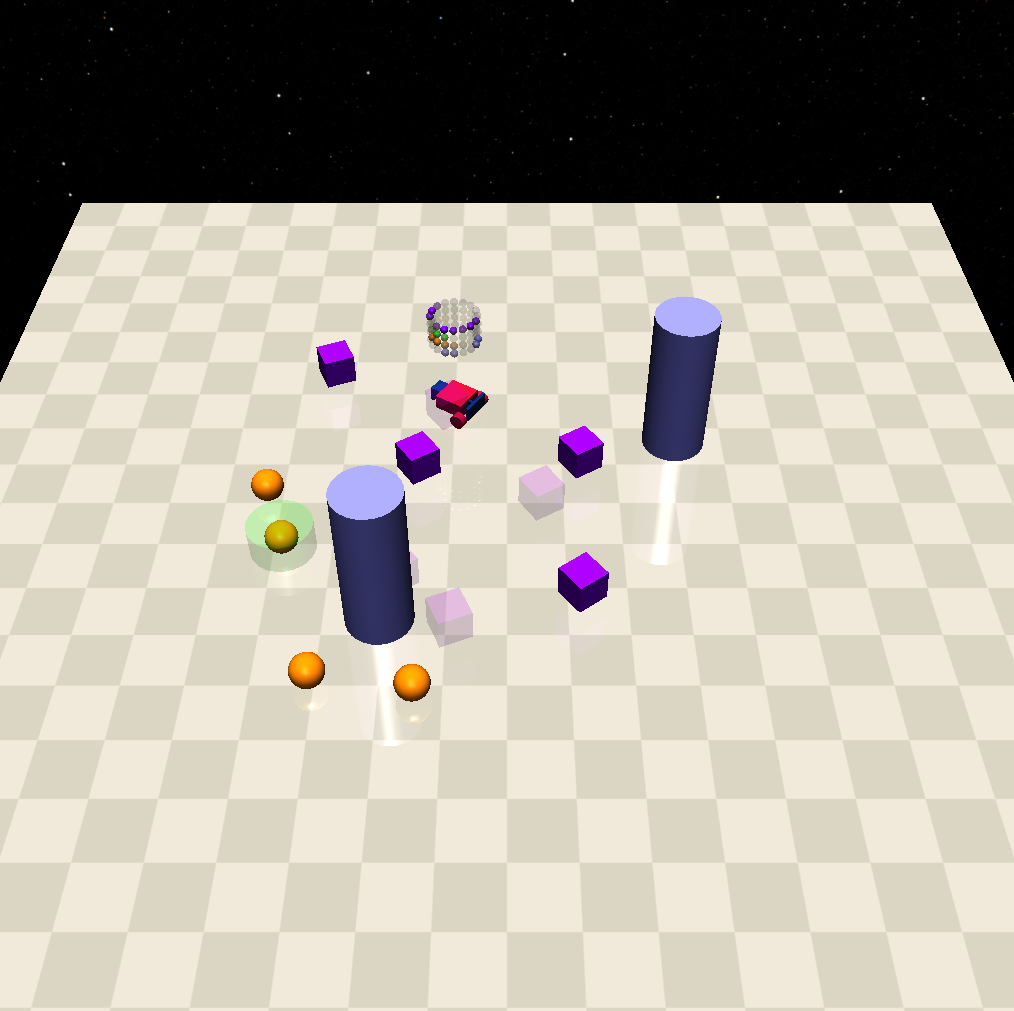}
    
    \smallskip
    \textbf{(c)} Randomised Button
\end{minipage}
\hfill
\begin{minipage}[t]{0.19\textwidth}
    \centering
    \includegraphics[width=\textwidth]{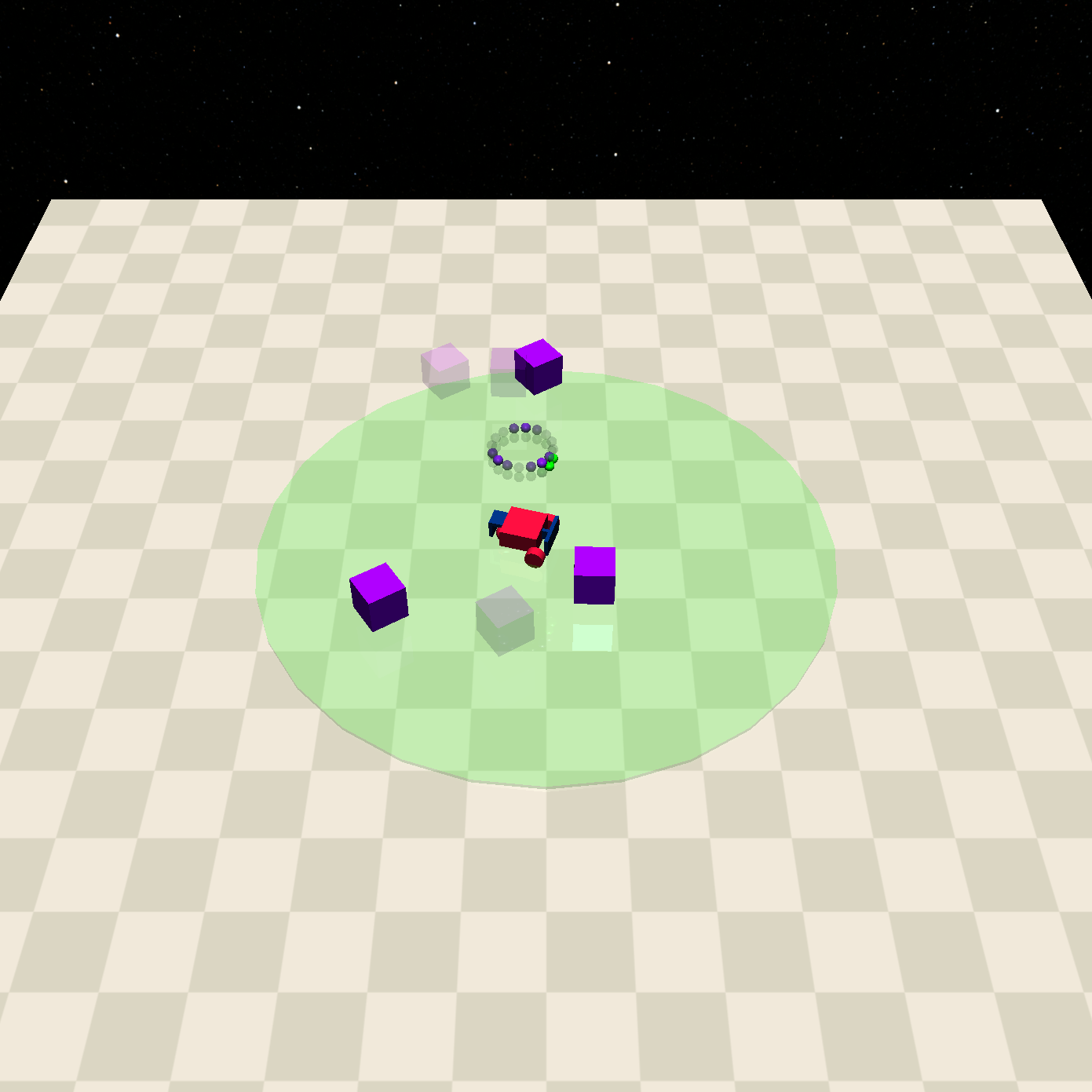}
    
    \smallskip
    \textbf{(d)} Randomised Circle
\end{minipage}
\hfill
\begin{minipage}[t]{0.19\textwidth}
    \centering
    \includegraphics[width=\textwidth]{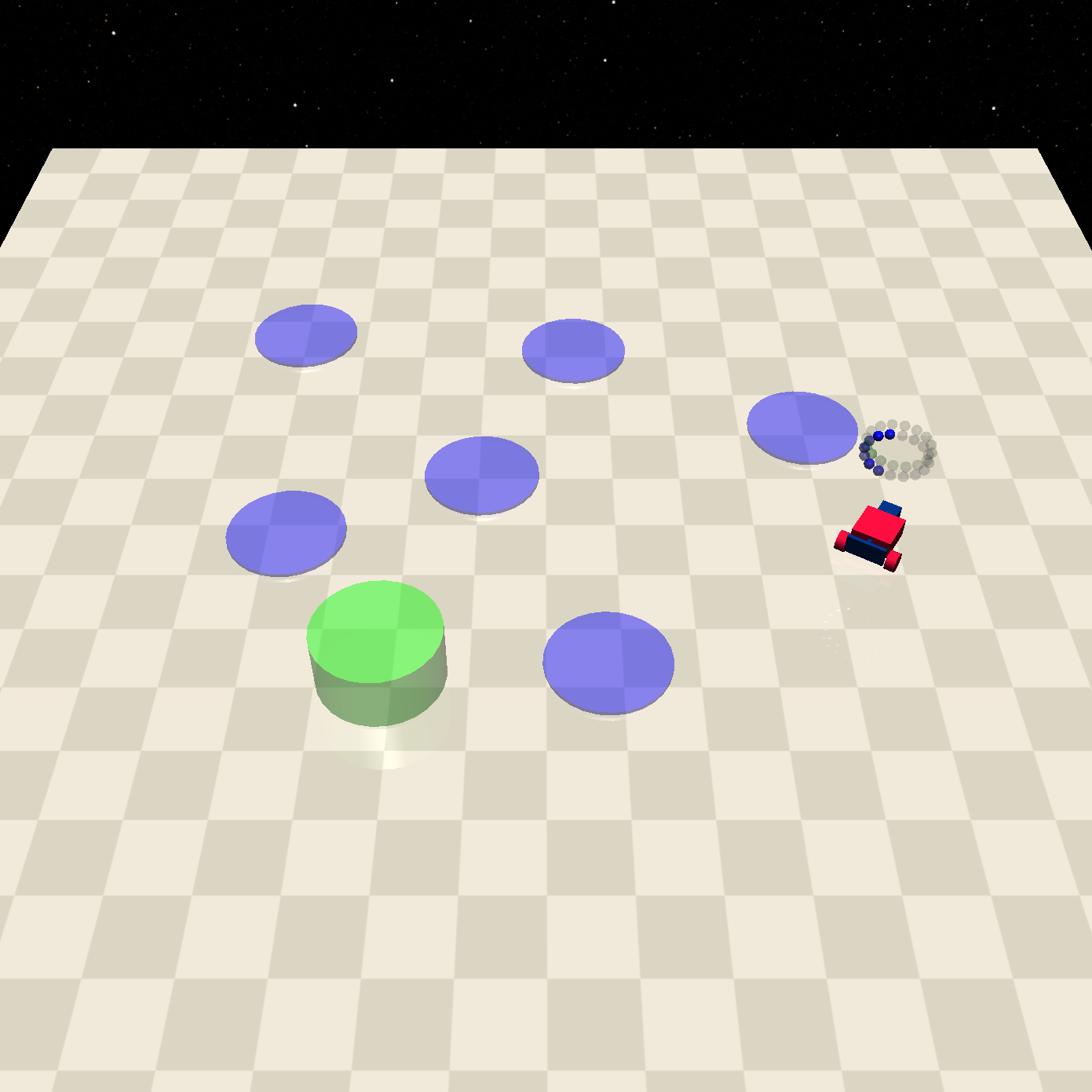}
    
    \smallskip
    \textbf{(e)} Constrained Randomised Goal
\end{minipage}
\caption{Example navigation tasks with varying complexity levels for evaluating \approach. The \textcolor{violet}{purple} blocks are the obstacles, the \textcolor{green}{green} circles are the goals, \textcolor{orange}{orange} spheres are the buttons, and the \textcolor{red}{red} vehicle is the RL agent.}
\label{fig:examples-envs}
\end{figure*}

\FloatBarrier


\section{Theoretical Analysis}
\label{sec:appendix-theo-analysis}
In this section, we theoretically analyze \approach\ and show that the expected probability of ADVICE misclassifying an unseen feature is bounded and can be decreased by diversifying the data collected before episode $E$.

\subsection{Theorem~\ref{thm:1}}

\textbf{Theorem 1.} 
The probability of \approach\ misclassifying a feature is bounded by $exp(-\gamma^2/2\sigma^2)$, where $\gamma$ is the contrastive separation margin and $\sigma^2$ is the variance of the assumed Gaussian noise in the latent space.

\begin{proof}
The contrastive separation margin in the latent space is given by:

\begin{equation}
\gamma = \min_{f_{s} \in F_{safe}, f_{u} \in F_{unsafe}} \| \textsc{Encode}(f_s) - \textsc{Encode}(f_u) \|_2
\label{eq:contrastive-seperation-margin}
\end{equation}

The noise in the latent space is assumed to follow a Gaussian distribution $\epsilon \sim N(0, \sigma^2)$. 
In ADVICE, an unseen feature $f$ is classified as safe if the K-nearest neighbours of $\textsc{Encode}(f)$ contain safer than unsafe features. 

Let 
$d_s = min_{f_s \in F_{safe}} \| \textsc{Encode}(f), \textsc{Encode}(f_s) \|_2$
and
$d_u = min_{f_u \in F_{unsafe}} \| \textsc{Encode}(f), \textsc{Encode}(f_s) \|_2$
define the Euclidean distance to the nearest safe and unsafe feature, respectively, given a new feature $f$.

An incorrect decision (misclassification) occurs when $f\in F_{safe}$ and $d_u < d_s$ or $f\in F_{unsafe}$ and $d_u > d_s$.
The contrastive separation margin $\gamma$ ensures that, in a noise-free case $\| \textsc{Encode}(f_u), \textsc{Encode}(f_s) \|_2 \geq \gamma \; \forall f_s \in F_{safe} \; \forall s_u \in F_{unsafe} $. 
In RL environments such as those targeted by \approach\ where noise exists, the distances $d_s$ and $d_u$ are perturbed by $\epsilon_s, \epsilon_u \sim N(0, \sigma^2)$. Therefore:

\ \\
\ \\

\begin{equation}
d_u - d_s = \| \textsc{Encode}(f), \textsc{Encode}(f_u) \|_2 - \| \textsc{Encode}(f), \textsc{Encode}(f_s) \|_2 \approx \gamma + \epsilon
\label{eq:perturbed-distances}
\end{equation}

where $\epsilon = \epsilon_u - \epsilon_s \sim N(0, \sigma^2)$ and is independent and Gaussian. 
The noise $\sigma$ in the latent space can come from: noisy data, imperfect model training, randomness in batch sampling, etc.
The probability of $d_u < d_s$ (misclassification) when $f\in S$ is given by:

\begin{equation}
P(d_u < d_s) = P(\gamma + \epsilon < 0) = P(\epsilon < -\gamma)
\label{eq:basic-misclassification}
\end{equation}

Since $\epsilon \sim N(0, \sigma^2)$, we can normalise it so that:

\begin{equation}
P(d_u < d_s) = P(Z < -\frac{\gamma}{\sigma})
\label{eq:normalised-misclassification}
\end{equation}

where $Z \sim N(0, 1)$. Using the cumulative distribution function of the standard normal distribution $\Phi$, we get:

\begin{equation}
P(d_u < d_s) = \Phi(-\frac{\gamma}{\sigma})
\label{eq:cdf-misclassification}
\end{equation}

\approach\ uses K-nearest neighbours to classify an unseen feature $f$. If $\gamma$ is large relative to $\sigma$, the probability of misclassifying an unseen feature decreases exponentially. So, we can define the expected probability of misclassifying a feature to be:

\begin{equation}
E[\textrm{feature misclassification}] \leq exp \left(-\frac{\gamma^2}{2\sigma^2} \right)
\label{eq:expected-misclassification}
\end{equation}
\end{proof}

\subsection{Proof of Theorem~\ref{thm:2}}

\textbf{Theorem 2.} 
The probability of \approach\ misclassifying a feature decreases exponentially with improved data entropy and is bounded by $exp(- \frac{H(F_E)}{2\sigma^2})$.

\begin{proof}
Let $\gamma_{m}$ express the effective achieved separation margin between sets $F_{safe}$ and $F_{unsafe}$, where $\gamma_{m} \leq \gamma$. Equality only holds under ideal conditions, such as perfectly diverse data, perfect model training, no data noise, etc. The diversity of the feature set $F_E$ collected before episode $E$ can be quantified using entropy:

\begin{equation}
H(F_E) = -\sum_{f\in F_E}p(f)\;log\;p(f)
\label{eq:data-entropy}
\end{equation}

where $p(f)$ is the probability distribution of features $f\in F_E$. Higher entropy corresponds to a broader set of features, ensuring greater diversity. Greater diversity results in more representative embeddings, given good model training, allowing the contrastive loss function to achieve better separation and clusterings of sets $F_{safe}$ and $F_{unsafe}$. The effective separation margin $\gamma_{m}$ depends on $H(F_E)$. As diversity increases, the embeddings for $F_{safe}$ and $F_{unsafe}$ become more separable, thereby $\gamma_{m} \;\propto\; k \cdot \sqrt{H(F_E)}$ where $k>0$ is a proportionality factor that links the entropy $H(F_E)$ of the feature set to the effective separation margin $\gamma_{m}$. It encapsulates the influence of latent space geometry, scaling properties, and model-specific parameters. While $k$ may vary depending on the training process and feature distribution, it is assumed to be stable for a given setup. Empirically, $k$ can be estimated by observing the relationship between $\gamma_{m}$ and $H(F_E)$ across diverse datasets or configurations. From Theorem 1, with the substitution of $\gamma_{m}$ for $\gamma$, the probability of misclassifying a feature is bounded by:

\begin{equation}
\begin{aligned}
P(misclassification) &\leq \exp\left(-\frac{\gamma_{m}^2}{2\sigma^2}\right) \\
                     &\leq \exp\left(-\frac{k\cdot H(F_E)}{2\sigma^2}\right)
\end{aligned}
\label{eq:gammaeff-miss-prob}
\end{equation}

therefore showing that increasing the diversity of the feature set reduces the misclassification probability of a feature exponentially.
\end{proof}

\subsection{Proof of Theorem~\ref{thm:3}}

\textbf{Theorem 3.} 
The expected regret of \approach\ with a fixed $K$ is upper bounded by a function of order $O\left(\frac{1}{\sqrt{E_{max}}}\right) + O\left(E_{max} \cdot \Delta Q \cdot exp\left(- \frac{\gamma^2}{2\sigma^2} \right) \cdot \frac{1}{\sqrt{K}} \right)$.

\begin{proof}
The total regret of \approach\ can be decomposed into two individual components: the baseline regret from the reinforcement learning process, and regret introduced from the shielding mechanism with a fixed $K$ accounting for misclassification. The baseline regret arises from the agent's standard exploration-exploitation trade-off in reinforcement learning~\citep{berry1985bandit}. This term is defined as:

\begin{equation}
= O \left(\frac{1}{\sqrt{E_{max}}} \right)
\label{eq:rl-regret}
\end{equation}

which reflects the improvement in the agent's policy quality over $E_{max}$ episodes. To construct the regret term for regret introduced from the shielding mechanism with a fixed $K$ accounting for misclassification, we can use the probabilistic bound of misclassification from Theorem 1, $exp\left(- \frac{\gamma^2}{2\sigma^2} \right)$. Let $\Delta Q = \sum_{t=1}^{T}\frac{Q^*(s_t, a_t) - Q^*(s_t, a^{'}_{t})}{T}$ denote the average regret per timestep caused by taking the misclassified action $a^{'}_{t}$ instead of the optimal action $a_t$. The shield also restricts the agent's action choices into a feasible set $\mathcal{F_K}$ determined by $K$ where: a higher $K$ (more cautious shield) reduces $\mathcal{F_K}$, whilst a lower $K$ (more relaxed shield) increases $\mathcal{F_K}$. Thus, the size of $\mathcal{F_K}$ scales inversely with $\sqrt{K}$. We can define our regret for the shielding term as:

\begin{equation}
= O \left(E_{max} \cdot \Delta Q \cdot exp\left(- \frac{\gamma^2}{2\sigma^2} \right) \cdot \frac{1}{\sqrt{K}} \right)
\label{eq:shield-regret}
\end{equation}
\end{proof}

\subsection{Proof of Theorem~\ref{thm:4}}

\textbf{Theorem 4.} 
The expected long-term regret of \approach\ with an adaptive $K$ is lower than \approach\ with a fixed $K$ and is upper bounded by a function of order $O\left(\frac{1}{\sqrt{E_{max}}}\right) + O \left(E_{max} \cdot \Delta Q \cdot exp\left(- \frac{\gamma^2}{2\sigma^2} \right) \cdot \frac{1}{\sqrt{K^*}} \right) + O \left(\frac{1}{E_{max}} \sum^{E_{max}}_{e=1} |K_e - K^* | \cdot \Delta Q_{K_{e}} \right)$.

\begin{proof}
The total regret of \approach\ with an adaptive $K$ can be decomposed into three individual components: the baseline regret from the reinforcement learning process, the regret introduced from the shielding mechanism with at the optimal $K^*$ accounting for misclassification, and the short-term regret incurred as $K$ deviates from the optimal $K^*$. We can use the baseline regret and the regret from the shielding mechanism from Theorem 3 with the substitution of $K^*$. The adaptive $K$ dynamically adjusts based on recent safety violations, thus leading to deviations from $K^*$. Therefore:

\begin{equation}
=  O \left(\frac{1}{E_{max}} \sum^{E_{max}}_{e=1} |K_e - K^* | \cdot \Delta Q_{K_{e}} \right)
\label{eq:adaptive-regret}
\end{equation}

reflects the cost of dynamically adjusting $K$ as the agent converges to $K^*$, where $|K_e - K^* |$ is the deviation of $K_e$ from $K^*$ at episode $e$ and $\Delta Q_{K_{e}}$ is the regret introduces by $K_e$ at episode $e$.

Next, we prove that the long-term regret of adaptive $K$ is lower than the fixed $K$. Adaptive $K$ adjusts dynamically based on the observed safety violations, this ensures that $K_e$ converges to an optimal $K^*$ as $E_{max} \rightarrow \infty$, thus minimizing the misclassification and shielding regret. Let $K^*$ minimize the combined regret:

\begin{equation}
K^* = argmin_{K} \left(\Delta Q \cdot exp\left(- \frac{\gamma^2}{2\sigma^2} \right) \cdot \frac{1}{\sqrt{K}} + \Delta Q_K \cdot \frac{1}{\sqrt{K}} \right)
\label{eq:combined-regret}
\end{equation}

where the first term represents misclassification regret which decreases with increasing K, and the second term represents shielding regret which increases with increasing K. The adaptive mechanism ensures that $K_e$ increases if safety violations are too frequent, and $K_e$ decreases if safety violations are rare. As $E_{max} \rightarrow \infty$, the mechanism stabilizes $K_e$ near $K^*$, leading to $|K_e - K^* | \rightarrow 0\; \text{as}\; e \rightarrow E_{max}$. The additional adjustment penalty from the adaptive $K$ mechanism (Equation~\ref{eq:adaptive-regret}) vanishes as $|K_e - K^* | \rightarrow 0$, leaving only:

\begin{equation}
= O\left(\frac{1}{\sqrt{E_{max}}}\right) + O \left(E_{max} \cdot \Delta Q \cdot exp\left(- \frac{\gamma^2}{2\sigma^2} \right) \cdot \frac{1}{\sqrt{K^*}} \right)
\label{eq:vanished-regret}
\end{equation}

For fixed $K$, the regret includes suboptimal shielding and misclassification effects: if $K > K^*$ then $K$ is too cautious and leads to high shielding regret, if $K < K^*$ then $K$ is too relaxed and leads to increased misclassification regret. Adaptive $K$, by converging to $K^*$, ensures that neither of these extremes persists in the long term, achieving a lower total regret.

While adaptive $K$ introduces early added regret, but diminishes as $K_e \rightarrow K^*$. This ensures that the long-term regret is dominated by the optimal $K^*$, unlike a fixed $K$, which does not adapt to the environment. Let $\epsilon$ represent the difference between $K$ and $K^*$. For a fixed $K$, $\epsilon$ is constant leading to regret terms proportional to $|\epsilon|$. For adaptive $K$, $|\epsilon_e|$ decreases over episodes due to convergence $|\epsilon_e| \propto \frac{1}{\sqrt{e}}$. This makes the additional regret introduced by the adaptive $K$ approach 0 as $E_{max} \rightarrow \infty$.
\end{proof}

\begin{figure*}[t]
\centering
\begin{minipage}{0.95\textwidth}
    \centering
    \includegraphics[width=\linewidth]{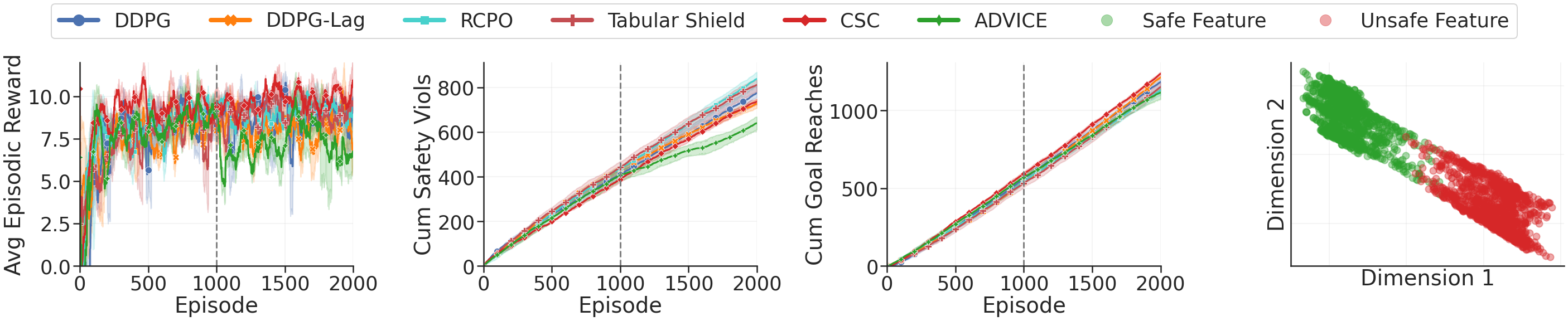}
    \smallskip
    \textbf{(a)} Full results on semi-random goal environment.
\end{minipage}
\begin{minipage}{0.95\textwidth}
    \centering
    \includegraphics[width=\linewidth]{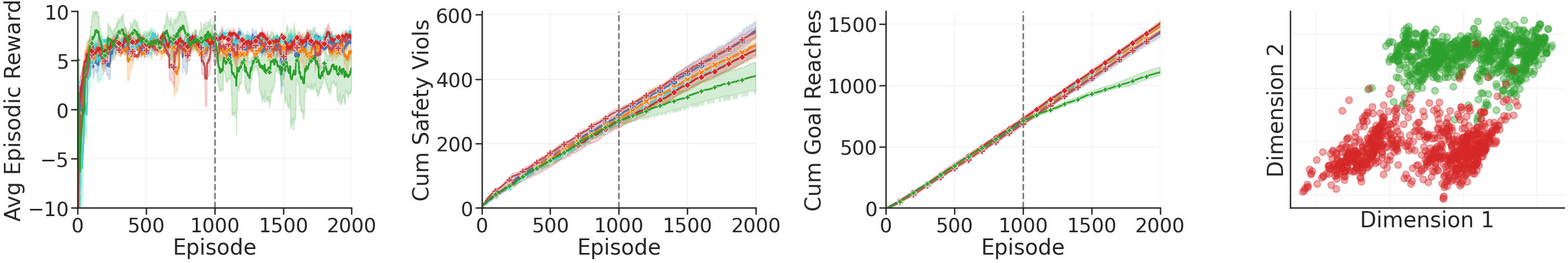}
    \smallskip
    \textbf{(b)} Full results on randomised goal environment.
\end{minipage}
\begin{minipage}{0.95\textwidth}
    \centering
    \includegraphics[width=\linewidth]{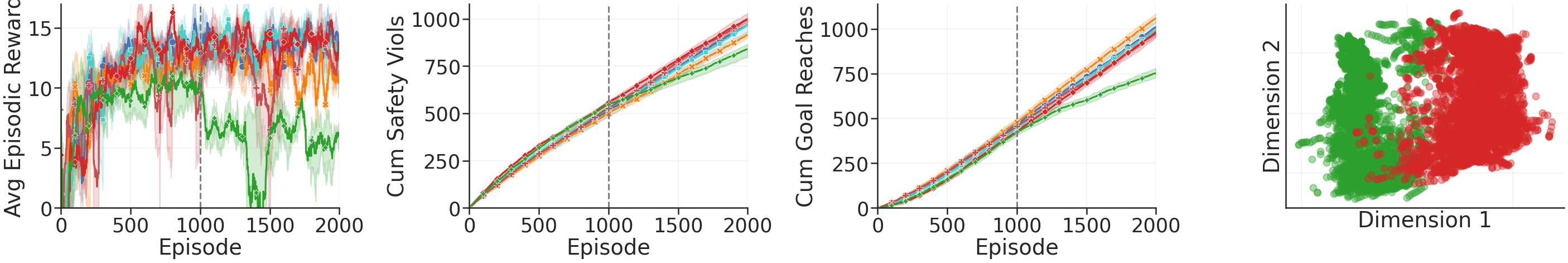}
    \smallskip
    \textbf{(c)} Full results on randomised button environment.
\end{minipage}
\begin{minipage}{0.95\textwidth}
    \centering
    \includegraphics[width=\linewidth]{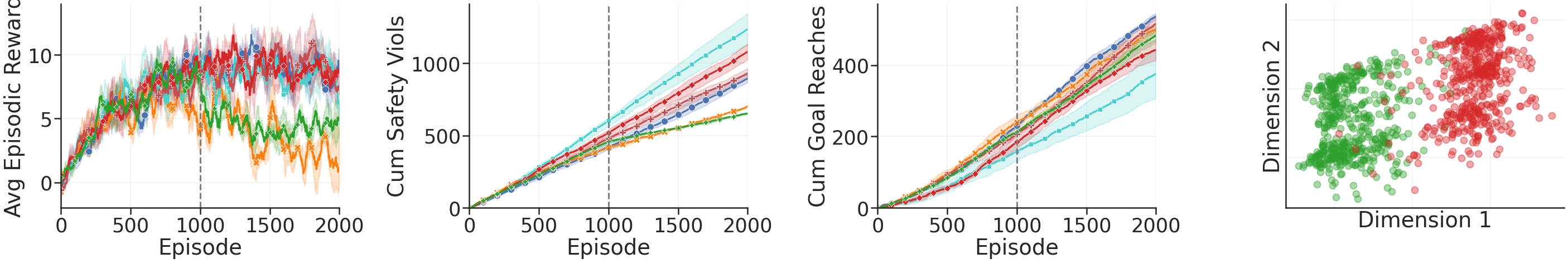}  
    \smallskip
    \textbf{(d)} Full results on randomised circle environment.
\end{minipage}
\caption{Average episodic reward, cumulative safety violations, cumulative goal reaches of examined approaches (DDPG, DDPG-Lag, RCPO, Tabular shield, Conservative Safety Critic, ADVICE) and example latent space visualisation for the semi-random goal (a), randomised goal (b), randomised button (c), and randomised circle (d) environments.}
\label{fig:appendix-full-results}
\end{figure*}


\section{Full Training Results}
\label{sec:appendix-full-results}

In Section~\ref{sec:core-results}, we show the main results for all approaches in a set of tasks. For fair comparison, we show results from episode $1000$ and standardise all metrics to zero. Below, in Figure~\ref{fig:appendix-full-results}, we show the unstandardised results for the same experiments.

\begin{figure*}[t]
    \centering
    \includegraphics[width=0.95\linewidth]{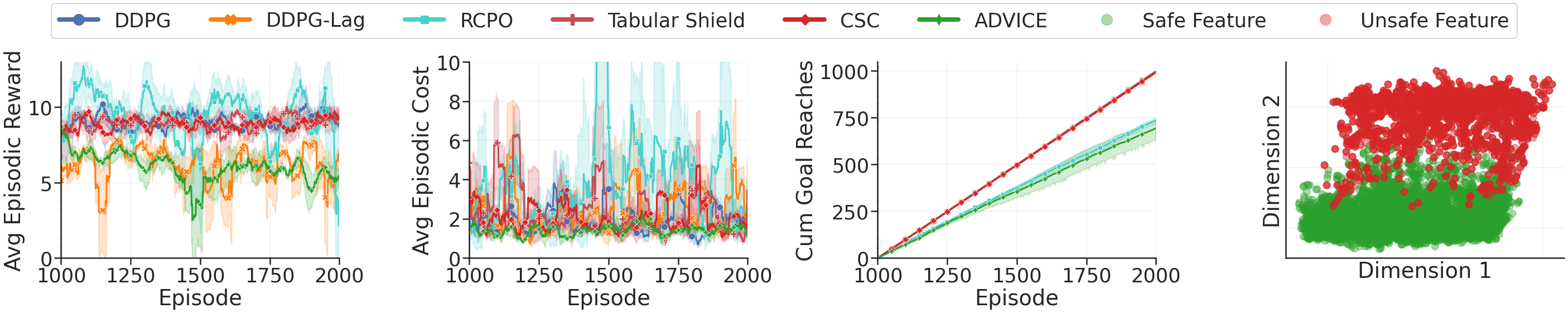}
    \caption{Average episodic reward, average episodic cost, cumulative goal reaches, and example latent space visualisation on the constrained randomised goal environment.}
    \label{fig:appendix-cmdp-results}
\end{figure*}

\begin{figure*}[t]
    \centering
    \includegraphics[width=0.95\linewidth]{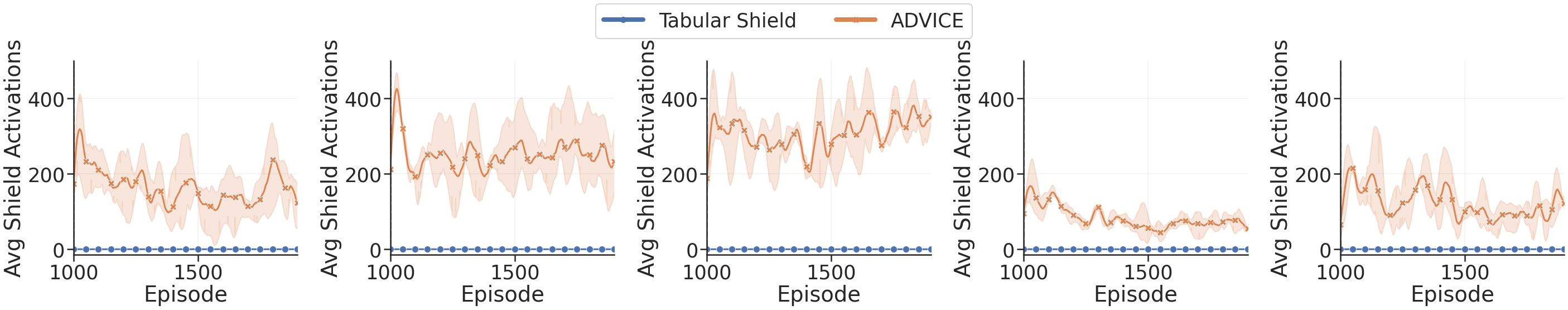}
    \caption{The average shield activations for \approach\ and the Tabular Shield in the semi-random goal, random goal, random button, random circle, and constrained random goal environments respectively.}
    \label{fig:appendix-shield-activations}
\end{figure*}

In all experiments, the Tabular Shield approach performs approximately the same as the standard DDPG agent. To show why this behaviour occurs, we plot the average shield activations for \approach\ and the Tabular Shield in Figure~\ref{fig:appendix-shield-activations}.
From these results, it is evident that the Tabular Shield does not activate once during training across all tasks. This is due to the high dimensionality of the environments evaluated. Even though the features stored are discretised to $1$ decimal place, the agent has to observe the exact same values across all $\approx 32$ dimensions plus the actions for the shield to activate. Our experiments show that this approach fails in these types of environments. A trend that can be noticed with \approach\ is when the shield is first activated, the amount of interventions starts relatively high. As training progresses, this number reduces, which shows that the agent learns to adapt to the shield's understanding of safety.

\begin{figure*}[t]
    \centering
    \includegraphics[width=0.90\linewidth]{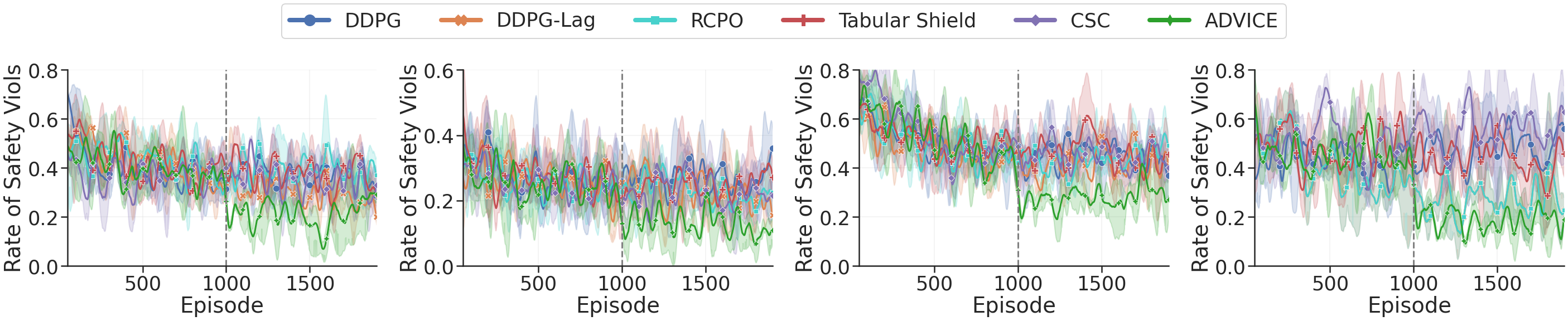}
    \caption{The rate of safety violations for all approaches in the semi-random goal, random goal, random button, and circle environments, respectively.}
    \label{fig:appendix-safety-viols}
\end{figure*}

Figure~\ref{fig:appendix-safety-viols} shows the rate of safety violations during training. This outcome further validates the results and conclusions discussed in Section~\ref{sec:core-results}. We observe that the DDPG and Tabular Shield agents perform similarly. The CSC agent, due to sparse data, underestimates safety and only reduces violations by a fractional amount. The DDPG-Lag agent manages to reduce the safety violations somewhat towards the end of training, which is particularly evident in the circle environment. Once \approach\ is turned on, it significantly reduces the rate of safety violations in all environments.

While constrained MDP environments are less common than unconstrained ones, we evaluated \approach\ in the Constrained Randomized Goal environment in Figure~\ref{fig:appendix-cmdp-results}. \approach\ consistently achieved one of the lowest average episodic costs, performing on par with cost-based approaches, which showed significant oscillations at times. These fluctuations highlight the difficulty these approaches face in balancing cost reduction with reward maximization~\citep{liu2022constrained}. Notably, such oscillations were absent in sparse-cost environments like the unconstrained settings in Section~\ref{sec:core-results}.


\section{Parameter Analysis}
\label{sec:appendix-other-tests}
In this section, we present an extended analysis of \approach\ to display the robustness and adaptability of the approach. These experiments were chosen to explore the effects of varying $K$ thresholds, and the timing of \approach's activations $E$.

\begin{figure*}[t]
    \centering
    \includegraphics[width=0.95\linewidth]{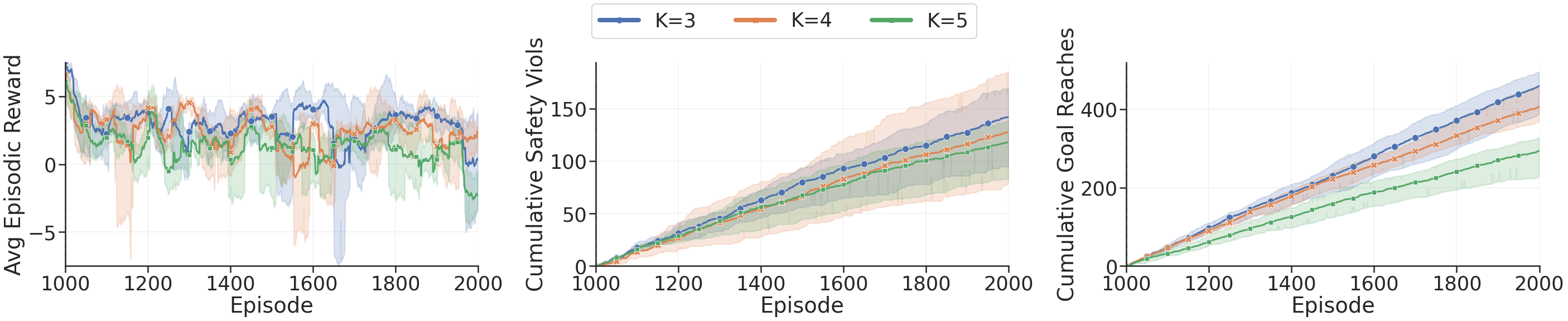}
    \caption{Average episodic reward, cumulative safety violations, and cumulative goal reaches of various values of $K$ on the randomised goal environment.}
    \label{fig:various-neighbours-results}
\end{figure*}

A user can specify the cautiousness of \approach\ using the safety threshold $K$. In Figure~\ref{fig:various-neighbours-results}, we evaluate how this parameter affects the model's safety, and performance. The results are clear, increasing $K$ leads to more cautious behaviour as hypothesised. The reward decreases a small amount as well as the cumulative goal reaches, however, it also results in fewer safety violations. Conversely, decreasing $K$ allows the underlying DDPG agent more freedom. As a result, average reward and goal reaches are increased at the expense of safety violations. These findings display a clear trade-off between return efficiency and safety assurance.

\begin{figure*}[t]
    \centering
    \includegraphics[width=0.75\linewidth]{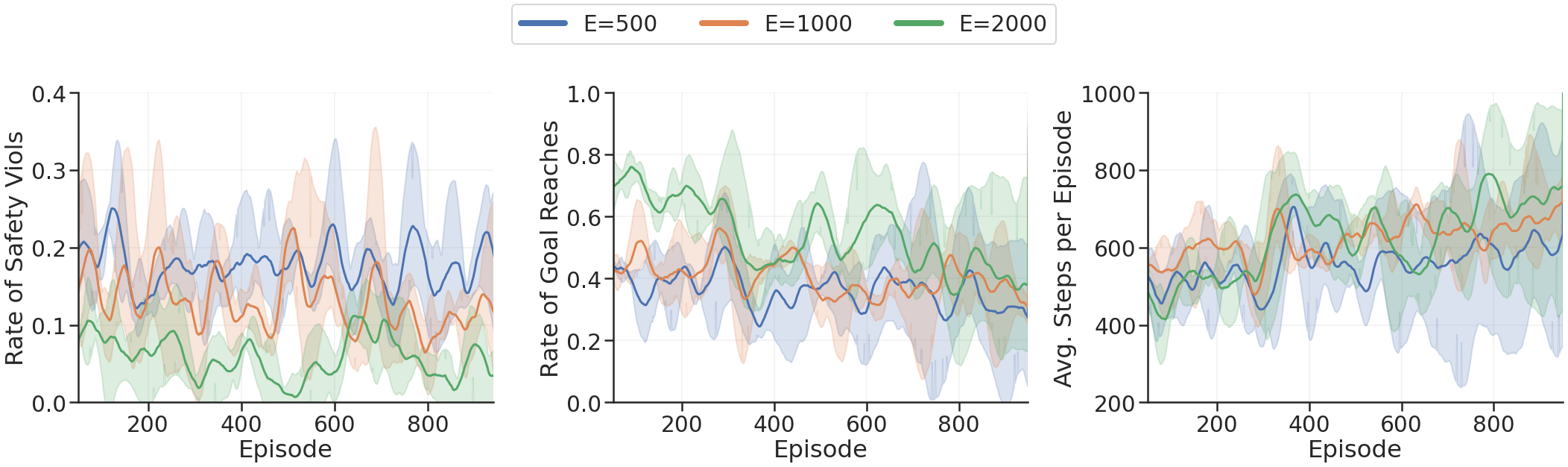}
    \caption{The rate of safety violations, and the rate of goal reaches when ADVICE is activated at different intervals $E$.}
    \label{fig:various-episode-results}
\end{figure*}

\approach\ has a \textit{cold-start}, meaning it requires some period of time before activation to collect data in order to work efficiently. We acknowledge that this can affect the performance of \approach\, evidently, we show the results of various activation points in Figure~\ref{fig:various-episode-results}. To allow for a fair comparison as possible, we show the rate of safety violations for the subsequent $1000$ episodes after activation. Again, we visualise a trade-off. Delaying \approach's activation for longer results in fewer safety violations and increased goal reaches. However, the RL agent observes more cumulative safety violations up to the point of activation. Starting \approach\ earlier decreases the number of safety violations up to activation, but gives the autoencoder fewer data points to train on. Evidently, safety violations are not reduced to the same magnitude, and goal reaches also decrease. This is to be expected with any neural network-based approach.

\begin{figure*}[t]
    \centering
    \includegraphics[width=0.95\linewidth]{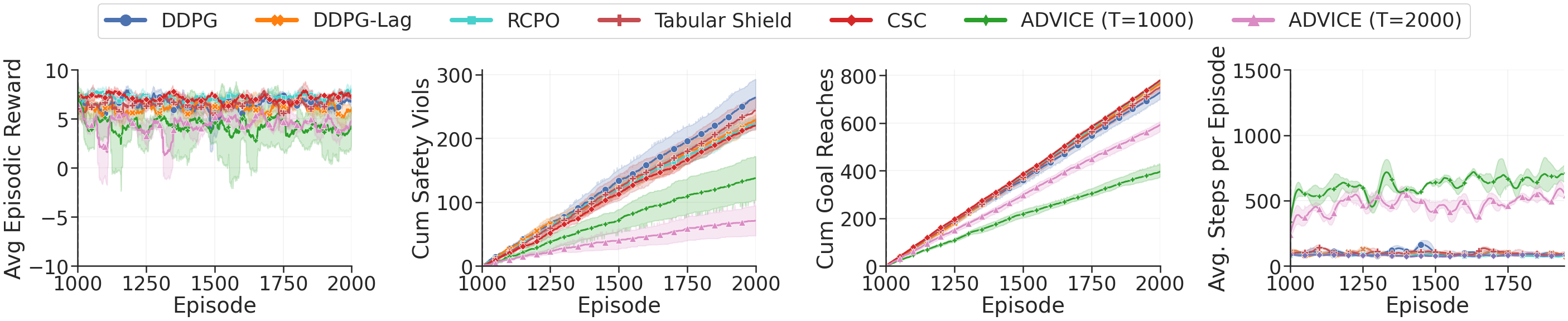}
    \caption{Average episodic reward, cumulative safety violations, and cumulative goal reaches of various approaches on the randomised goal environment where the maximum episodic steps are doubled.}
    \label{fig:various-steps-results-appendix}
\end{figure*}

Based on results in Figure~\ref{fig:core-results}b, we hypothesise that the reduced reward and cumulative goal reaches is a result of \approach\ not having \textit{enough} time to complete the task. As seen in Figure~\ref{fig:example-trajectories}, \approach\ learns to take a longer route to the objective, so by doubling the maximum step count allowed per episode, we expect to see an increase in cumulative goal reaches, average reward, and no increase in safety violations. Results for this experiment are shown in Figure~\ref{fig:various-steps-results}. As expected, given more time to complete the task, \approach\ now reaches the goal more than when the maximum step counter is the default value. As a result, we observe an increase in average episodic return much closer to baseline approaches.


\section{Adaptive ADVICE}
\label{sec:appendix-adaptive}

To evaluate \approach's adaptation capabilities, we conducted a sensitivity analysis focusing on the impact of varying the distant time window $h_d$ and the recent time window $h_r$. 
For each $h_d$ and $h_r$ combination, Table~\ref{tab:adaptive-shield-results} shows the number of consecutive episodes $K$ was fixed at a specific value, the frequency of $K$ adjustments, and the impact on mean safety violations. 

Our findings reveal an impact of $h_r$ on the frequency of $K$ adjustments. 
A smaller $h_r$ leads to more frequent $K$ increases, allowing the system to quickly respond to immediate safety violations. Conversely, a larger $h_r$ tends to stabilise $K$ by filtering out anomalies and adjusting only in response to sustained trends of increased violations. Similarly, $h_d$ influences the decrease of $K$; a smaller $h_d$ facilitates rapid decreases in $K$ following a reduction in safety violations, whereas a larger $h_d$ results in less frequent reductions, promoting stability in \approach's behaviour.
The interaction between $h_d$ and $h_r$ minimally affects the overall rate of safety violations, suggesting that while these parameters impact the adaptiveness and stability of $K$, they do not directly correlate with safety violations. These insights highlight the role of $h_d$ and $h_r$ primarily as tuning parameters to balance responsiveness against stability in \approach.

\begin{table*}[htb]
\centering
\sffamily
\caption{Mean sensitivity analysis of $h_d$ and $h_r$ on the randomised goal environment for $E^{max}\!=\!1000$}
\label{tab:adaptive-shield-results}
\begin{small}
\begin{tabular}{@{}llll@{}}
\toprule
\multicolumn{2}{l}{}                            & \multicolumn{2}{c}{$h_{d}$} \\ \cmidrule(l){3-4} 
\multicolumn{1}{c}{$h_{r}$} & \multicolumn{1}{c}{Metrics} & \multicolumn{1}{c}{10} & \multicolumn{1}{c}{25} \\ \midrule
\multirow{5}{*}{2} & Consecutive Episodes ($K = 3$)  & $24.49 \pm 9.59$  & $31.09 \pm 4.53$  \\
                    & Consecutive Episodes  ($K = 4$) & $1.00 \pm 0.00$   & $7.89 \pm 0.77$   \\
                    & Consecutive Episodes  ($K = 5$) & $32.97 \pm 7.95$  & $151.60 \pm 3.09$ \\
                    & Changes of $K$                  & $116.33 \pm 9.87$ & $27.67 \pm 4.62$  \\
\multirow{-5}{*}{2} & Safety Violations               & $50.42 \pm 9.94$  & $57.88 \pm 7.64$  \\ \midrule
                    & Consecutive Episodes  ($K = 3$) & $166.81 \pm 5.71$ & $117.35 \pm 8.07$ \\
                    & Consecutive Episodes  ($K = 4$) & $21.10 \pm 6.86$  & $28.76 \pm 9.26$  \\
                    & Consecutive Episodes  ($K = 5$) & $13.49 \pm 5.40$  & $149.68 \pm 8.71$ \\
                    & Changes of $K$                  & $28.67 \pm 1.41$  & $19.33 \pm 2.83$  \\
\multirow{-5}{*}{3} & Safety Violations               & $51.69 \pm 6.93$  & $54.45 \pm 5.10$  \\ \midrule
                    & Consecutive Episodes  ($K = 3$) & $954.49 \pm 9.59$ & $169.07 \pm 9.02$      \\
                    & Consecutive Episodes  ($K = 4$) & $22.60 \pm 6.71$  & $22.10 \pm 7.26$  \\
                    & Consecutive Episodes  ($K = 5$) & $0.00 \pm 0.00$   & $181.35 \pm 4.92$ \\
                    & Changes of $K$                  & $2.33 \pm 0.47$   & $17.77 \pm 2.72$  \\
\multirow{-5}{*}{4} & Safety Violations               & $55.48 \pm 5.64$  & $50.17 \pm 5.47$  \\ \bottomrule
\end{tabular}
\end{small}
\end{table*}

\begin{figure*}[t]
    \centering
    \includegraphics[width=0.75\linewidth]{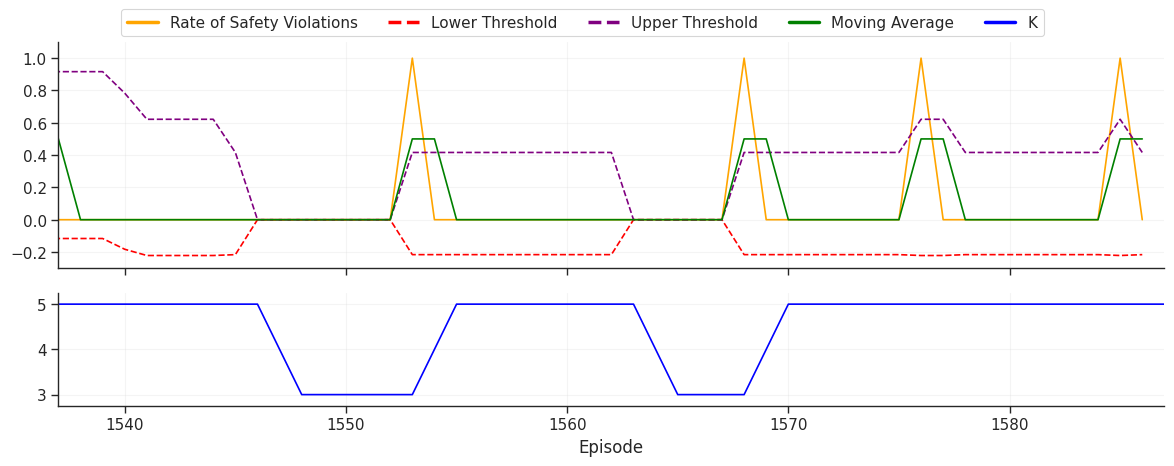}
    \caption{The rate of safety violations, lower threshold $(\mu_{d} - \sigma_{d})$, upper threshold $(\mu_{d} + \sigma_{d})$, recent moving average $(\mu_{r})$, and value of $K$ during an example run where $h_d=10$ and $h_r=2$.}
    \label{fig:appendix-ten_two_example}
\end{figure*}

To further validate that Adaptive ADVICE correctly increases and lowers $K$ during training, we plot an example visualisation window in Figure~\ref{fig:appendix-ten_two_example} showing the rate of safety violations, the upper $(\mu_{d} + \sigma_{d})$ and lower $(\mu_{d} - \sigma_{d})$ thresholds, the moving average $(\mu_{r})$, and the value of $K$.

It can be seen that when the recent moving average $(\mu_{r})$ is above the upper threshold $(\mu_{d} + \sigma_{d})$, the adaptive module correctly increments $K$. An example of this can be seen at episode $1553$. The agent crashes, and both thresholds adjust but the recent moving average climbs above the upper threshold, increasing $K$ as a result. In subsequent episodes afterwards, the recent moving average falls between both thresholds. Here the adaptive module correctly keeps $K$ at the same value until episode $1564$ where the moving average is equal to the lower threshold. As a result, $K$ is decreased. This example window validates that the adaptive shield works as expected and also provides an insight into how it works during training.

\begin{figure*}[t]
    \centering
    \includegraphics[width=0.75\linewidth]{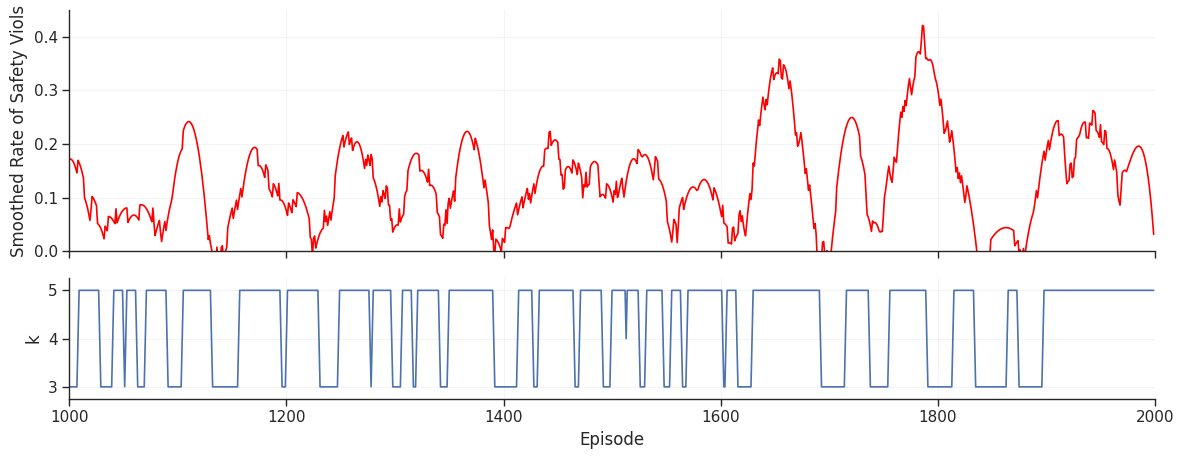}
    \caption{Rate of safety violations and value of $K$ on an example run (one random seed) that shows the adaptation of ADVICE ($h_d = 10, h_r = 2$).}
    \label{fig:appendix-ten_two_example-full}
\end{figure*}


\section{Hyperparameter Analysis and Computational Overheads}
\label{sec:appendix-hyperparams}

This section lists the hyperparameters used by all models and ADVICE. Table~\ref{tab:hyperparameter-configuration} summarises all hyperparameters used in Section~\ref{sec:evaluation}. We will refer the reader to our source code repository for the remaining details.

Using all model configurations in Table~\ref{tab:hyperparameter-configuration}, a single ADVICE run (one random seed) takes 12, 24, 24, 12, and 12 hours of training, respectively, in the semi-random goal, random goal, random button, random circle, and constrained random goal environments. For all other approaches, a single run takes 3, 4, 5, 3, and 5 hours in the same environments, respectively. All experiments were run on a large computing cluster utilising two Nvidia H100 GPUs, 16 CPUs, and up to 500GB memory.

\begin{table}[t]
\vspace{4mm}
\centering
\captionsetup{justification=centering}
\caption{Summary of hyperparameters in the DDPG algorithm and the ADVICE shield.}
\label{tab:hyperparameter-configuration}
\begin{tabular}{@{}lllll@{}}
\toprule
Parameter & DDPG &  & Parameter & ADVICE Shield \\ \midrule
Network size & (256, 256) &  & Size of network & (512, 2, 512) \\
Optimizer & Adam &  & Optimizer & NAdam \\
Actor learning rate & 2e-3 &  & Learning rate & Reduce on plateau \\
Critic learning rate & 1e-3 &  & Batch size & 32 \\
Size of replay buffer & 2e5 &  & Max epochs & 500 \\
Batch size & 64 &  & No. Neighbours ($K_{max}$) & 5 \\
Gamma & 0.95 &  & No. Safe neighbours ($K$) & 4 \\
Tau & 5e-3 &  & Losses & (MSE, MSE, CL) \\
Ornstein-Uhlenbeck noise & 0.2 &  & Loss weights & (1, 1, 1.25) \\
- & - & & Unshielded Episodes ($E$) & 1000 \\ \bottomrule
\end{tabular}
\end{table}

Hyperparameters for the DDPG algorithm started with author recommendations~\citep{ddpg-paper}. They were manually tuned afterwards to achieve a high performance on individual environments before tests were carried out, meaning the RL algorithm for all approaches was of high performance and fair comparison. Hyperparameters for \approach\ were manually tuned for performance in CL loss and MSE loss. Some hyperparameter analysis was conducted in Section~\ref{sec:appendix-other-tests} to justify certain choices. Parameters for the DDPG-Lag approach started with recommendations~\citep{stooke2020responsive} and were tuned for performance in our experiments.


\end{document}